\documentclass[a4paper,fleqn]{cas-sc}

\usepackage[numbers]{natbib}
\usepackage{lineno}
\usepackage{hyperref}
\usepackage{textcomp}
\DeclareUnicodeCharacter{3B1}{\ensuremath{\alpha}}
\usepackage{float}
\usepackage{subfig} 
\usepackage{overpic} 
\usepackage[ruled,linesnumbered]{algorithm2e}
\usepackage{multirow}
\usepackage{makecell}
\usepackage{ulem}
\usepackage{graphicx}
\usepackage{color}

\def\tsc#1{\csdef{#1}{\textsc{\lowercase{#1}}\xspace}}
\tsc{WGM}
\tsc{QE}
\tsc{EP}
\tsc{PMS}
\tsc{BEC}
\tsc{DE}

\begin{document}

\let\WriteBookmarks\relax
\def\floatpagepagefraction{1}
\def\textpagefraction{.001}
\shorttitle{CAT: A Causal Graph Attention Network for Trimming Heterophilic Graphs}
\shortauthors{Silu He, Qinyao Luo, Xinsha Fu, Ling Zhao, Ronghua Du, Haifeng Li. CAT: A causal graph attention network for trimming heterophilic graphs. Information Sciences. 2024, 677, 120916.}

\title [mode = title]{CAT: A Causal Graph Attention Network for Trimming Heterophilic Graphs}

\author[1]{Silu He}
\author[1]{Qinyao Luo}
\author[2]{Xinsha Fu}
\author[1]{Ling Zhao}
\author[3]{Ronghua Du}
\author[1]{Haifeng Li}
\cormark[1]

\cortext[cor1]{Corresponding author: Haifeng Li, Email: lihaifeng@csu.edu.cn}
\address[1]{School of Geosciences and Info-Physics, Central South University, Changsha 410083, China}
\address[2]{School of Civil Engineering and Transportation, South China University of Technology, Guangzhou 510640, China.}
\address[3]{College of Automotive and Mechanical Engineering, Changsha University of Science and Technology, Changsha 410114, China.}

\begin{abstract}
The local attention-guided message passing mechanism (LAMP) adopted in graph attention networks (GATs) can adaptively learn the importance of neighboring nodes and perform local aggregation better, thus demonstrating a stronger discrimination ability. However, existing GATs suffer from significant discrimination ability degradations in heterophilic graphs. The reason is that a high proportion of dissimilar neighbors can weaken the self-attention of the central node, resulting in the central node deviating from its similar nodes in the representation space. This type of influence caused by neighboring nodes is referred to as Distraction Effect (DE) in this paper. To estimate and weaken the DE induced by neighboring nodes, we propose a Causal graph Attention network for Trimming heterophilic graphs (CAT). To estimate the DE, since DE is generated through two paths, we adopt the total effect as the metric for estimating DE; To weaken the DE, we identify the neighbors with the highest DE (we call them Distraction Neighbors) and remove them. We adopt three representative GATs as the base model within the proposed CAT framework and conduct experiments on seven heterophilic datasets of three different sizes. Comparative experiments show that CAT can improve the node classification accuracies of all base GAT models. Ablation experiments and visualization further validate the enhanced discrimination ability of CATs. In addition, CAT is a plug-and-play framework and can be introduced to any LAMP-driven GAT because it learns a trimmed graph in the attention-learning stage, instead of modifying the model architecture or globally searching for new neighbors. The source code is available at https://github.com/GeoX-Lab/CAT.

\end{abstract}




\maketitle
\section{Introduction}

Graph neural networks (GNNs) are the most reliable and prevailing benchmark models for graph learning. With their effectiveness at representing irregular graph data, GNNs achieve state-of-the-art performance in tasks such as node classification, link prediction, graph classification, graph generation, and graph similarity calculation. They have also been widely applied in various fields such as recommendation systems, computer vision, natural language processing, molecular, and transportation. Their graph representation capability primarily stems from the ability to aggregate information  \cite{update1}, essentially following the message passing mechanism, which can build invariant input representations for the central node based on its neighbors. Existing GNNs utilize various aggregation operations following their fundamental assumptions about the influence of neighbors. However, they are all founded on the strong homophily hypothesis, obeying the rule that neighbors tend to be similar \cite{19}. Among these GNNs, the graph attention network (GAT) \cite{20} is a representative network that adaptively learns the importance of neighbors for aggregation through the local attention-guided message passing Mechanism (LAMP); therefore, it has the potential to achieve better performance on high-homophily graphs. However, the reverse of this situation is that the GATs' performance decreases when addressing low-homophily graphs because assigning different aggregation weights under the smoothing principle leads to the failure to aggregate beneficial information and disrupt the raw features \cite{add16}. Experiments have shown that GNNs exhibit significant declines in node classification tasks \cite{19} when the input graph is heterophilic, and we find that LAMP-driven GATs exhibit the most notable declines (as shown in Section \ref{sec4.1}). The primary reason for this phenomenon is the high proportion of dissimilar neighbors. Dissimilar neighbors influence the representation of the central node through their assigned attention levels and the weakened self-attention of the central node; both situations can result in the central node deviating from its similar nodes in the representation space. We refer to this impact of neighboring nodes on the central node as Distraction Effect (DE), which is generated through two paths (capturing the attention assigned to neighbors and reducing the self-attention of the central node). Improving the discrimination capabilities of GATs on heterophilic graphs poses a significant challenge.

Great efforts have been devoted to improving the discrimination ability of general GNNs on heterophilic graphs, with only a few offering specific solutions for GATs. Based on the fundamental strategy targeted at GNN models or input graph data, these approaches can be categorized into two groups: GNN architecture-based tactics and graph structure-based tactics. \textbf{GNN architecture-based tactics} methods focus on modifying the GNN architecture to better utilize the information from neighboring nodes for aggregation. gNovel aggregation mechanisms have been proposed to adjust the weights of neighbors \cite{25,24,23,21,22,26,add16}, and some works have fused information derived from different GNN layers in a new way \cite{28,30,29,27,add12,update6}. Self-supervised learning has also been adopted to capture more information from neighbors \cite{add17,31,update5,32,add10,add11,add14,33,add13,add17}. Some methods focus on improving attention mechanisms \cite{43,42,update7}. \textbf{Graph structure-based tactics} methods, on the other way, focus on making heterophilic graphs more homophilic by obtaining more similar nodes for aggregation. These methods search for high-order neighbors \cite{35,34,add15,update4} or nearer neighbors in latent spaces \cite{37,39,40,36,41,38}, forcing the central node closer to similar nodes in the representation space.

In general, existing methods primarily focus on addressing a single issue: how to enhance the process of aggregating information from other nodes? GNN architecture-based tactics target at the \textbf{aggregation mechanism}, while graph structure-based tactics target at the \textbf{aggregation source}. The former involves the weights assignment, feature transformation, and learning paradigms for aggregation; while the latter proposes strategies for selecting the set of nodes for aggregation. They represent two different perspectives on improving \textbf{aggregation} respectively. However, the aggregation operation is derived from the strong homophily hypothesis, which is not satisfied by heterophilic graphs. Therefore, modifying the aggregation operation is not essential. In addition, searching for aggregation sources with higher similarity is not necessary for explaining the poor performance of GNNs on heterophilic graphs \cite{19} and has the potential to cause oversmoothing.

\begin{figure}
	\centering
	\includegraphics[width=3in]{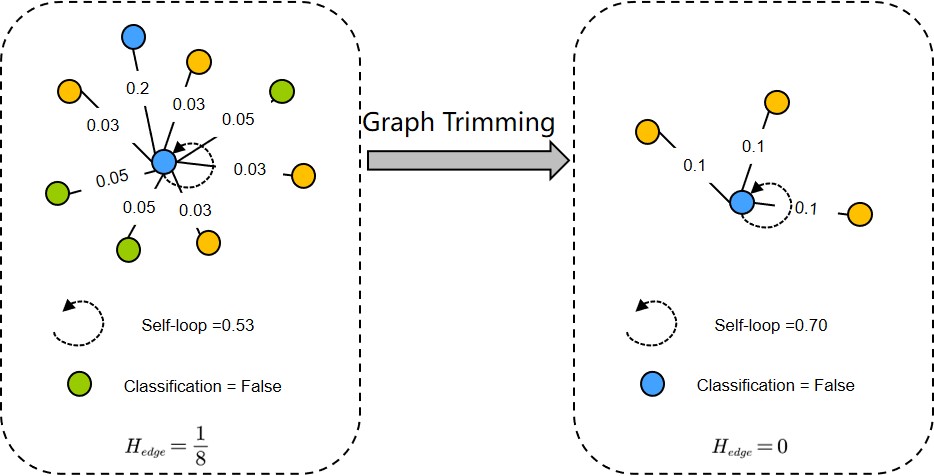}
	\caption{Toy example: A comparison between scenarios with low self-attention and high self-attention for the central node.}
	\label{fig1}
\end{figure}

Contrary to the emphasis on aggregation, we propose a new insight concerning the mechanism of GATs: \textbf{enabling the central node to concentrate on itself and avoiding the distraction during the aggregation can improve the discrimination ability of GATs on heterophilic graphs.} We illustrate a representative example in Figure \ref{fig1}. For heterophilic graphs, a high proportion of interclass edges leads to the updated representation of the central node deviating from the distribution of its class, even when similar neighbors are assigned higher weights. After graph trimming, despite the decreased homophilic ratio, the representation of the central node deviates less and is classified correctly due to the higher self-attention and lower distraction level. Since the removed nodes contribute to the decreased self-attention of the central node, and removing them helps prevent deviations, we refer to these nodes as Distraction Neighbors. They are mathematically equal to the neighbors with high DEs.

To identify and remove Distraction Neighbors, we need to measure the DE of neighboring nodes on the central node, that is, the effect of the neighbors on the attention distribution of the central node. Therefore, two crucial questions must be answered.

\textbf{Question 1:} What is the basic unit of Distraction Neighbors when influencing the attention-learning of the central node?

\textbf{Answer 1:} Using two heterophilic graphs as an example, we intervene in the local neighbor distribution (LND) of the nodes and obtain three control groups to explore the effect of the LND on the discrimination ability of the central node (Section \ref{sec4.1}). Experiments reveal that nodes belonging to the same class provide similar semantic information; this kind of information is referred to as the Class-Level Semantic. Based on this observation, we introduce the concept of Class-Level Semantic Cluster and further propose the \hyperref[Hyp1]{Class-Level Semantic Space hypothesis} in (Section \ref{sec4.2}). According to this hypothesis, we believe that neighbors belonging to the same class have similar impacts on identifying the central node; therefore, the basic unit for measuring DE should be the class. It is more beneficial to obtain genuine and stable effects of neighbors by treating the neighbors belonging to the same class as a group.

\textbf{Module 1.} Based on Answer 1, we design a \hyperref[module1]{Class-level Semantic Clustering Module}, to precluster local neighbors and obtain different Semantic Clusters for measuring their DE on the central node.

\textbf{Question 2:} To what extent do the Distraction Neighbors influence the attention-learning of the central node?

\textbf{Answer 2:} To better estimate the DE, we model the DE as a type of causal effect. Specifically, we formalize the influencing paths of neighboring nodes on the attention-learning process of the central node based on the working mechanism of the GAT and construct causal graphs (Figure \ref{fig2} and Figure \ref{fig8}). Since the neighboring nodes influence the central node through two paths, we chose the total effect to estimate the overall causal effect.

\textbf{Module2.} Based on Answer 2, we design a \hyperref[module2]{Total Effect Estimation Module}, to intervene in the LND of central nodes with Semantic Cluster as the basic unit, and then calculate the TE from the changes in the attention distribution of the central node before and after the intervention. Distraction Neighbors are identified and removed according to the TE, and a corresponding trimmed graph is generated.

Our contributions are as follows:
\begin{enumerate}
\item We propose a novel insight for enhancing the discrimination ability of GATs on heterophilic graphs: maintaining the self-attention of the central node and avoiding distraction caused by neighbors. Instead of altering the architecture of the GAT or searching for new neighbors globally, we use the attention distribution learned by GAT to identify and remove Distraction Neighbors, which can be regarded as performing a trimming operation on the graph.
\item We propose a Causal graph Attention network for Trimming heterophilic graphs (CAT), to improve the discrimination ability of GATs for heterophilic graphs. We employ three GATs as the base model and conduct node classification experiments on seven datasets of three sizes. Comparison experiments, ablation experiments, and visualization experiments validate the effectiveness of CAT.
\item We conduct pre-experiments and investigate the mechanism by which the LND influences the attention-learning of the central node based on our observations and background knowledge. We further formalize this idea into causal graphs. 
\end{enumerate}

The remainder of this paper is organized as follows: in Section \ref{sec2}, we classify and summarize existing GNN methods for heterophilic graphs. In Section \ref{sec3}, we introduce important concepts and background knowledge needed for this paper, including the causal graphs derived from the background knowledge; In Section \ref{sec4}, we present the pre-experiments and the hypotheses we drew from them. We introduce our method in Section \ref{sec5} and describe the dataset and experiments in Section \ref{sec6}. In Section \ref{sec7} and Section \ref{sec8}, we discuss and conclude this work, fundamental issues that need further investigation are also raised.

\section{Related Work\label{sec2}}
The strong homophily assumption underlying graphs indicates that connected nodes are similar, which is a necessity of GNNs. This principle is also widely acknowledged in various domains such as social networks and citation networks. Under this assumption, aggregating the information of neighbors gradually brings nodes belonging to the same class closer in the representation space, thereby improving the discrimination ability of GNNs. However, when confronted with heterophilic graphs, the merits of GNNs may not be realized. The declines exhibited by GATs are particularly pronounced (Section \ref{sec4.1}). GNNs for heterophilic graphs have attracted increasing attention, and we categorize the approaches aimed at overcoming these challenges into two groups based on their fundamental strategies: GNN architecture-based tactics and graph structure-based tactics.

\textbf{GNN architecture-based tactics.} The fundamental question addressed by methods in this line is how to more effectively aggregate information from neighbors. Therefore, these methods design and build various GNN architectures to better learn and fuse the information of neighbors.

\begin{enumerate}
    \item \textbf{Some methods aim to modify the aggregation operation in message passing. }Various kinds of graph information are leveraged to guide the neighbor propagation process, where aggregation weights are learned to enhance similar features and weaken dissimilar features.
    An ordered GNN \cite{25} leverages a rooted-tree hierarchy aligning strategy to order message passing, thereby achieving better fusing of information provided by nodes in different hops. NHGCN \cite{24} employs a new metric, Neighborhood Homophily (NH) to group and aggregate the neighbors differently. LW-GCN \cite{23} proposes a labelwise message passing mechanism that uses pseudolabels to guide the aggregation of similar nodes and preserve heterophilic contexts. DMP \cite{21} takes attributes as weak labels to measure the attribute homophily rate, and to specify the attribute weights of the edges for aggregation. CPGNN \cite{22} incorporates an interpretable compatibility matrix for modelling the heterophily or homophily level, and uses this matrix to propagate and update the prior belief of each node. GGCN \cite{26} proposes two strategies, structure-based and feature-based edge correction to adjust the edge weights for aggregation. SAGNN \cite{add16} implements a sign attention mechanism to adaptively learn the weights of neighbors, which aggregates positive and negative information for neighbors within the same class and in different classes, respectively.
    
    \item \textbf{Some methods aim to design different GNN layers and determine their relationships.} Because different layers in a GNN can encode different levels of node features, specific information can be learned by combining different intermediate layers. 
    Auto-HeG \cite{28} builds a comprehensive GNN search space from which the optimal heterophilic GNN is selected. IIE-GNN \cite{30} designs a GNN framework that contains seven blocks in four layers to enrich the intraclass information extraction process. H2GCN \cite{29} uses a combination of intermediate layers to concatenate the node representations derived from all previous layers, thereby better capturing local and global information. GPR-GNN \cite{27} combines a Generalized PageRank algorithm with a GNN to learn the weights of GNN layers for combination with the intermediate layer representation. PCNet \cite{add12} employs a PC-Conv to perform both homophilic and heterophilic aggregation of node information, and SPCNet \cite{update6} further improves this approach.
    
    \item  \textbf{Some methods aim to train GNNs in new learning paradigms. }Self-supervised learning is a new paradigm that can help models learn better representations by leveraging the intrinsic structure of graphs. Multiview learning is a popular method that learns from multiple views via contrastive learning or invariance regularization \cite{add17}, to capture rich information from unlabelled nodes.
    HLCL \cite{31} and PolyGCL \cite{update5} use graph filters to generate augmented graph views and contrast the high-pass filter representation with the low-pass representation for conducting graph contrastive learning under heterophily. MVGE \cite{32} builds two augmentation views with input ego features and aggregated features, and forces the model to learn different graph signals through a graph reconstruction task. GREET \cite{add10} trains an edge discriminator to augment homophilic and heterophilic views, and then uses a dual-channel contrastive loss to learn node representations. LHS \cite{add11} adopts a self-expressive generator to induce a latent homophilic structure via multinode interactions and iteratively refines the latent structure with a dual-view contrastive learner. MUSE \cite{add14} performs cross-view feature fusion across semantic and contextual views and learns perturbation-invariant representations via contrastive learning. SimP-GCN \cite{33} employs a contrastive pretext task to capture the complex similar and dissimilar feature relations between nodes, which can help the method conduct node similarity-preserving aggregation. A multiresolution graph contrastive learning method \cite{add13} has been proposed that learns resolution invariant representations from graph augmentations constructed by diffusion wavelet filters. HGRL \cite{add17} adopts four types of graph augmentations and two pretext tasks to capture graph properties.

    \item \textbf{Some methods focus on GAT solutions,}  which are referred to as \textbf{GAT-oriented methods}. These approaches consider the characteristics of the GAT and perform better aggregation by proposing novel attention mechanisms.
     HA-GAT \cite{43} utilizes a heterophily-aware attention scheme to adaptively assign weights for edges, and learns the local attention pattern of the central node by learning the importance of each heterophilic edge type. GATv3 \cite{42} implements a new attention architecture to compute the attention coefficients between the query and key, which optimizes the representations of nodes by introducing representations learned by other GNNs. DGAT \cite{update7} leverages the diffusion distance to detect noisy neighbors and rewires heterophilic graphs, and proposes global directional attention to capture long-range neighborhood information.
\end{enumerate}

\textbf{Graph structure-based tactics. }The core question behind this type of approach is how to select the neighbors that can provide beneficial information for aggregation. Therefore, these methods primarily involve restructuring a meaningful graph to connect more similar neighbors and then aggregating their information.

\begin{enumerate}
    \item \textbf{Some methods seek similar neighbors from high-order neighbors.} With the experimental observation \cite{35} that high-order neighborhoods may have higher homophily ratios, aggregating information from higher-order neighborhoods can lead to satisfactory performance.
    U-GCN \cite{35} uses a multitype convolution mechanism to capture and fuse the information from 1-hop, 2-hop, and kNN neighbors. GPNN \cite{34} adds the most relevant nodes from a large number of multihop neighborhoods, and filters out irrelevant or noisy nodes from the local neighborhoods. PathMLP \cite{add15} designs a similarity-based path sampling strategy guided by hop-by-hop similarity to conduct homophilic path aggregation. SFA-HGNN \cite{update4} uses high-order random walks to select and aggregate distant nodes.
    
    \item \textbf{Some methods search for nearest neighbors in a learned feature space.} Graph representation learning methods aim to embed graphs into a latent space that approximates the inherent distribution space of the input data as closely as possible. Therefore, neighbors in this latent space contribute to better central node representations.
    GEOM-GCN \cite{37} learns a latent space and aggregates information from neighbors in the latent space to strengthen the ability of GCN to capture long-range dependencies in heterophilic graphs. Non-local GNN \cite{39} leverages attention mechanisms to sort and find distant but informative nodes for conducting nonlocal aggregation. HOG-GCN \cite{40} designs a novel propagation mechanism guided by the homophily degree between node pairs learned in the homophily degree matrix estimation module. GCN-SL \cite{36} uses spectral clustering to construct a reconnected graph according to the similarities between nodes and performs aggregation on it. A graph restructuring method \cite{41} based on adaptive spectral clustering improves the node classification accuracy of GNNs by improving graph homophily. DHGR \cite{38} rewires the graph by adding homophilic edges and pruning heterophilic edges, and the similarity of label/feature distribution of node neighbors is adopted to determine the rewiring strategy.
\end{enumerate}

Our approach differs from the above GNNs for heterophilic graphs in that it does not require alterations of the original GNN models or global searching for new neighbors, but instead removes Distraction Neighbors via graph trimming. We make full use of the attention distribution learned by the original GAT models as signals, to find a better attention distribution. Therefore, our method is plug-and-play and is applicable to any LAMP-driven GAT.

\section{Preliminaries\label{sec3}}
\textbf{Semi-supervised Graph Node Classification.} Graph node classification is a fundamental task in graph representation learning, to classify graph nodes into predefined categories \cite{48}, and can be used as a proxy task for measuring the discrimination ability of graph representation models. Existing methods mainly focus on the semi-supervised paradigm. Given a graph $G=(V, E)$, where $V$ is the set of nodes and $E$ is the set of edges. $A \in \mathbb{R}^{N \times N}$ is the adjacency matrix of the graph, and $X \in \mathbb{R}^{N \times F}$ is the node feature matrix. The number of layers in the GNN model is $K$, and the node representation in layer $k \in\{1,2, \ldots K\}$ is $z^k \in \mathbb{R}^{N \times H}$, where $H$ denotes the representation dimension in the hidden layer. In this task, each node belongs to a specific category, only the labels of the nodes in the training set are visible, and the goal is to predict the category of unlabelled nodes. 

\textbf{Graph Attention Network.} GAT is a graph neural network architecture that can adaptively learn the importance of neighboring nodes by leveraging an attention mechanism to obtain the weights of neighbors \cite{20}. A graph attention layer depicts how to obtain the representation $z^k \in \mathbb{R}^{N \times H_k}$ in layer $k$ with the input of representation  $z^{k-1} \in \mathbb{R}^{N \times H_{k-1}}$ in layer $k-1$. The attention coefficient $\alpha_{i j}$ between a pair of nodes $(i, j)$ where $ A_{i j} \neq 0$ is calculated by a linear transformation layer $W \in \mathbb{R}^{H_{k-1} \times H_k}$  and a shared attention mechanism $a: \mathbb{R}^{H_{k-1}} \times \mathbb{R}^{H_k} \rightarrow \mathbb{R}$ according to Eq.\ref{equal1}.

\begin{equation}
\label{equal1}
\alpha_{i j}=\frac{\exp \left(\sigma\left(W_2\left[W Z_i^{k-1}|| W Z_j^{k-1}\right]\right)\right)}{\sum_{m \in N(i)} \exp \left(\sigma\left(W_2\left[W Z_i^{k-1}|| W Z_m^{k-1}\right]\right)\right)}
\end{equation}

Let $H_i^k=\sigma\left(\sum_{j \in N(i)} \alpha_{i j} W Z_j^{k-1}\right)$ denote the updated representation of the central node $V_i$. $W_2 \in \mathbb{R}^{1 \times 2 H^k}$ is a linear transformation layer, $\sigma(\cdot)$ is a nonlinear activation function, and $\|$ represents the concatenation operation. If a multihead attention mechanism is applied, a node representation is calculated for each attention head, and the final representation is calculated with all attention heads according to Eq.\ref{equal2}.

\begin{equation}
\label{equal2}
H_i^k=\Delta_{t=1}^T H_i^k
\end{equation}

Here $\Delta(\cdot)$ stands for concatenation, averaging or another pooling operation.

\textbf{Causal Inference.\label{CI}} Causal inference is a new data science \cite{49}, that involves making causal claims rather than merely associational claims based on the belief that causality is inherently more stable. It is concerned with (1) causal discovery (Is there a causal relationship between two variables? How does the cause impact the effect?) and (2) causal effect estimation (How much does the cause impact the effect?) Important notations used in this paper are as follows.

\begin{itemize}
    \item \textbf{Cause and Effect.} A variable $X$ is identified as a cause of a variable $Y$ if $Y$ can change in response to changes in $X$. Alternatively, we can say that $Y$ is 'Listen to' $X$. Then $X$ is the cause and $Y$ is the effect. If $Y$ directly responds to $X$, then $X$ is the direct cause of $Y$. 
    \item \textbf{Causal Graph.} A causal graph is a Directed Acycling Graph (DAG) that models the causality with graphical language. In causal graphs, every parent is a direct cause of its children.
    \item \textbf{Intervention.} If we intervene in a variable $Z$ in a causal graph, it deletes all the edges from its parent variables and sets the intervened variable to $\bar{z}$. We can denote this operation as $do(Z=\bar{z})$. The children of $Z$ change naturally with the change in $Z$.
    \item \textbf{Total Effect (TE).\label{TE}} Total effect measures the whole effect of $X$ on $Y$, including the direct effect and indirect effect. The TE can be calculated by $TE_{x \rightarrow \bar{x}}=Y(X=x)-Y(do(X=\bar{x}))$.
\end{itemize}

\begin{figure}
	\centering
	\includegraphics[width=4in]{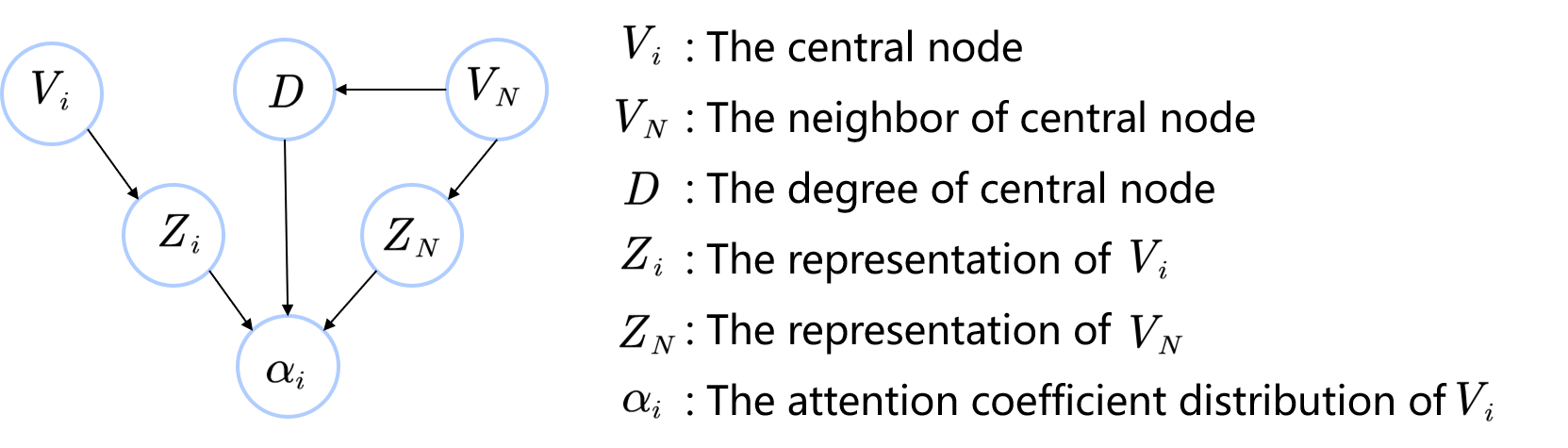}
	\caption{Causal graph behind GAT.}
	\label{fig2}
\end{figure}

With the preliminaries we stated above, we construct the causal graph underlying GATs in accordance with Eq.\ref{equal1} and Eq.\ref{equal2}, as shown in Figure \ref{fig2}.

\begin{itemize}
    \item $V_i \rightarrow Z_i \rightarrow \alpha_i \leftarrow Z_N \leftarrow V_N$: The attention coefficient distribution of the central node $V_i$ is calculated from the representation of $V_i$ and $V_N$.
    \item $V_N \rightarrow D \rightarrow \alpha_i$: When the attention coefficients are normalized, the neighboring nodes influence the attention distribution of the central node through the degree of the central node.
\end{itemize}

Notably, $V_N$ affects the final attention distribution $α_i$ of $V_i$ through two causal paths. On the one hand, the representation of $V_N$ affects its importance to $V_i$. On the other hand, the degree of $V_i$ changes due to the existence of $V_j \in V_N$, thereby influencing the final attention coefficient distribution when normalizing $\alpha$. To measure the effect of one (or more) neighboring node(s) on the learned attention of the central node, we choose TE to calculate the causal effect of neighboring nodes, which is adopted as measurements of their DE.

\begin{figure}
	\centering
	\includegraphics[width=3in]{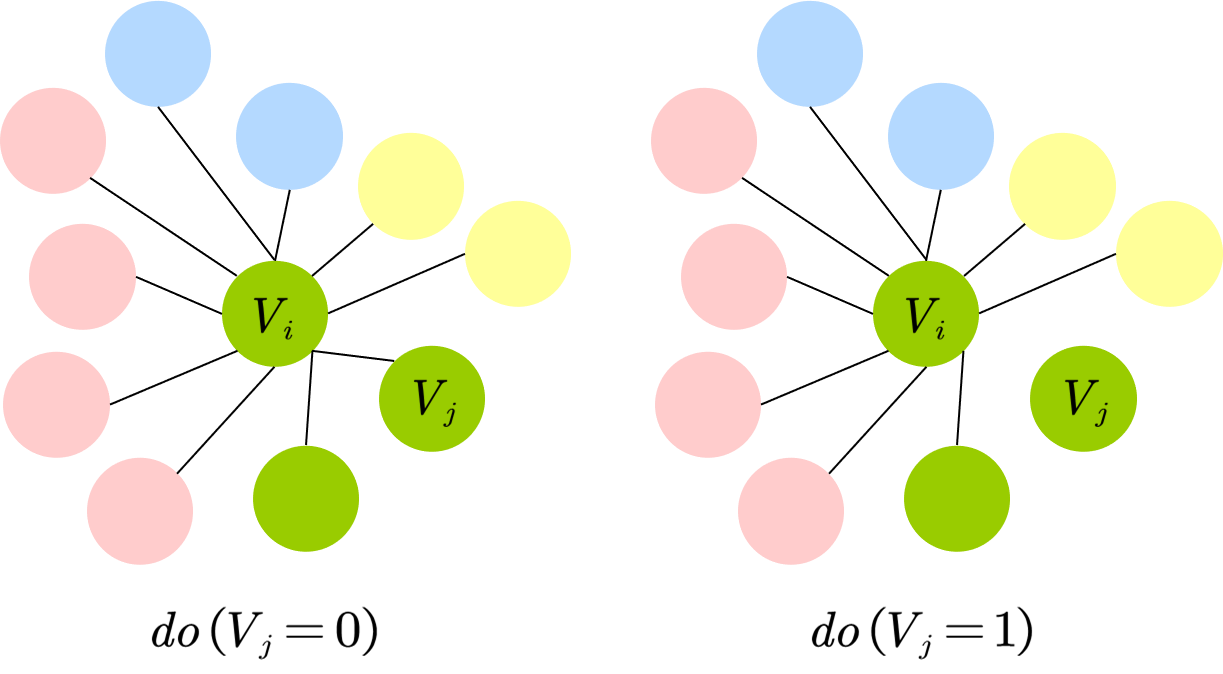}
	\caption{Intervention implemented for the local neighbor distribution of the central node.}
	\label{fig3}
\end{figure}

We estimate the TE by intervening in the LND of $V_i$. As illustrated in Figure \ref{fig3}, for a neighboring node $V_j \in V_N$, $V_j=0$ represents the reservation of $V_j$ as a neighbor of $V_i$, while $V_j=1$ represents the removal of  $V_j$ from $V_N$. According to Eq.\ref{equal3}, we can estimate the effect of $V_j$ on the attention coefficient distribution of $V_i$.

\begin{equation}
\label{equal3}
TE_{\alpha_i}=E_{\alpha_i \mid d o\left(V_j=1\right)}\left[\alpha_i \mid \operatorname{do}\left(V_j=1\right)\right] - E_{\alpha_i \mid d o\left(V_j=0\right)}\left[\alpha_i \mid \operatorname{do}\left(V_j=0\right)\right]
\end{equation}

Similarly, we denote the self-attention coefficient that $V_j$ assigns to itself as $\alpha_{self\_attention}$, and we can obtain the TE of $V_j$ on the self-attention of $V_i$ according to Eq.\ref{equal4}:

\begin{equation}
\label{equal4}
\begin{split}
    TE_{\alpha_{self\_attention}} & =E_{\alpha_{self\_attention} \mid d o\left(V_j=1\right)}\left[\alpha_{self\_attention } \mid \operatorname{do}\left(V_j=1\right)\right] \\
    &- E_{\alpha_{self\_attention} \mid d o\left(V_j=0\right)}\left[\alpha_{self\_attention } \mid \operatorname{do}\left(V_j=0\right)\right]
\end{split}
\end{equation}

\section{Pre-experiments and Key Hypothesis\label{sec4}}

In this section, we illustrate our observations derived from the pre-experiments designed in Section \ref{sec4.1} and propose the key hypothesis in Section \ref{sec4.2}. In the pre-experiments, we disentangle the effects of neighboring nodes into two factors and intervene in them to generate different intervention graphs as treatment groups. The experimental results indicate that nodes in the same class can provide similar semantic information for discrimination, where \hyperref[Hyp1]{Class-level Semantic Space Hypothesis} can be derived. We also propose the inference of Class-level Semantic Space Hypothesis, \hyperref[Inf1]{Low Distraction and High Self-attention}, which is the core strategy of our method. 

\subsection{The Effect of the Local Neighbor Distribution(LND)\label{sec4.1}}

\begin{figure}[htbp]
	\centering
	\subfloat[]{\includegraphics[width=.2\columnwidth]{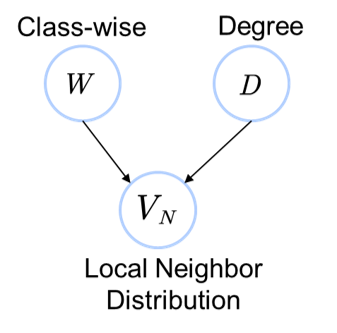}\label{fig4a}}\hspace{5pt}
	\subfloat[]{\includegraphics[width=.4\columnwidth]{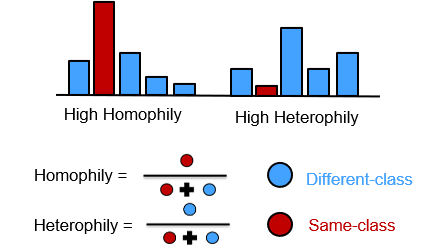}\label{fig4b}}
	\subfloat[]{\includegraphics[width=.3\columnwidth]{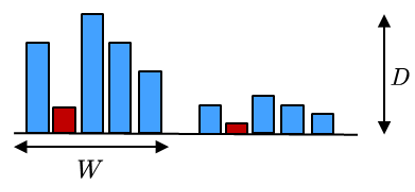}\label{fig4c}}\hspace{5pt}
	\caption{Local Neighbor Distribution (LND). (a) The two factors that influence LND. (b) How Class-wise influences the LND and the homophily of graph. (c) How Degree influences the LND.}
\end{figure}

GNNs are renowned for the ability to aggregate the information provided by neighboring nodes and update the representation of the central node. Therefore, the local neighbor distribution (LND) is an important contributing factor to the ability of GNN models. As illustrated in Figure \ref{fig4a}, the LND can be decomposed into two factors, Class-wise ($W$) and Degree ($D$). The former statistically characterizes the distribution of neighboring nodes in different classes, and the latter denotes the number of neighboring nodes. Figure \ref{fig4b} illustrates that different $W$ will change the local homophily of the central node, and Figure \ref{fig4c} illustrates that LNDs with the same homophily are significantly different under different $D$. We formalize the LND of nodes in graph $G$ as $LND_G=\left\{W_c, D_c\right\}, c \in C$, where $C$ denotes the set of node classes in $G$.

\begin{figure}
	\centering
	\includegraphics[width=6in]{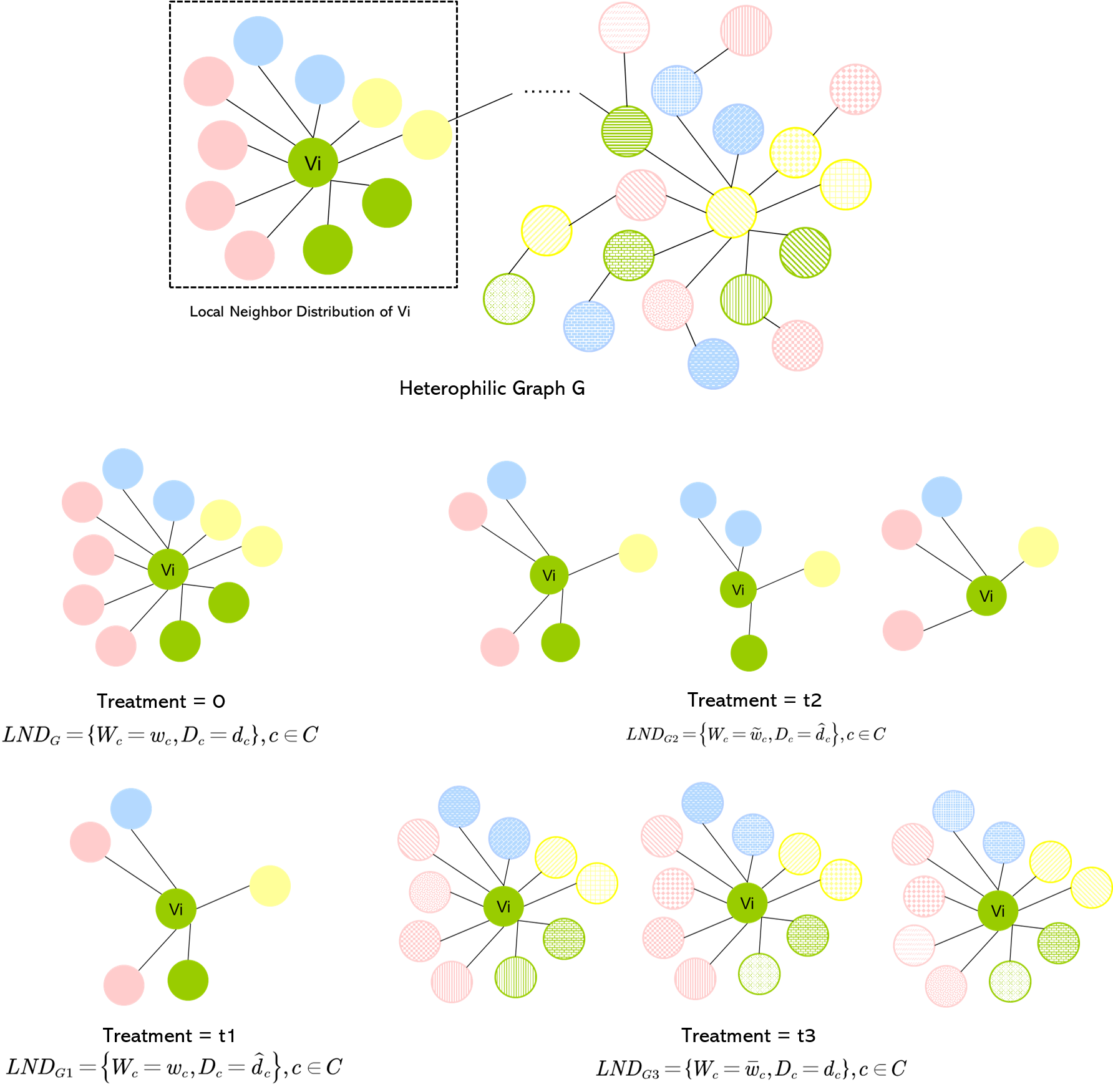}
	\caption{The control and treatment group settings used in the pre-experiment.}
	\label{fig5}
\end{figure}

To further determine the influence of the LND on the discrimination ability of GNNs, we used $G$ as a control group and intervened in the LND of $G$ to construct different treatment groups. Then, we conduct control experiments on three representative GNN models, GCN \cite{50}, GraphSAGE \cite{51} and GAT \cite{20}, and compare their node classification accuracies (the outcomes, which are represented as $Y$) on different groups. The experimental settings are illustrated in Figure \ref{fig5} and Table \ref{tab1}. We choose two heterophilic graph datasets, Chameleon and Squirrel to conduct the pre-experiment. The experimental settings are as follows:
\begin{enumerate}
    \item \textbf{Control group:} Original graph $G$.
    \item \textbf{Treatment Group 1:} The $D$ of the central node decreases, and the $W$ remains constant.
    \item \textbf{Treatment Group 2:} The $D$ of the central node decreases, and the $W$ is set randomly. We set three random groups with random seeds of 0, 10, and 100.
    \item \textbf{Treatment Group 3:} The $D$ of the central nodes remains constant, while the neighboring nodes are randomly replaced with different nodes belonging to the same class. We search for replacement nodes with random seeds 0, 10, and 100.
\end{enumerate}

\begin{table} [width=1\linewidth,cols=4,pos=ht]
\caption{Settings for control experiments. }
\label{tab1}
\begin{tabular}{cc} 
\toprule
Graph/Group  & LND \\ 
\midrule
$G$ /treatment=0 & $LND_G=\left\{W_c=w_c, D_c=d_c\right\}, c \in C$  \\
$G_1$ /treatment=$t_1$  & $LND_{G1}=\left\{W_c=w_c, D_c=\hat{d}_c\right\}, c\in C$  \\ 
$G_2$ /treatment=$t_2$   &$LND_{G2}=\left\{W_c=\widetilde{w}_c, D_c=\widehat{d}_c\right\}, c \in C$\\
$G_3$ /treatment=$t_3$   & $LND_{G3}=\left\{W_c=\bar{w}_c, D_c=d_c\right\}, c \in C$  \\
\bottomrule
\end{tabular}
\end{table}

\begin{figure}
	\centering
	\subfloat[Chameleon]{\includegraphics[width=.45\columnwidth]{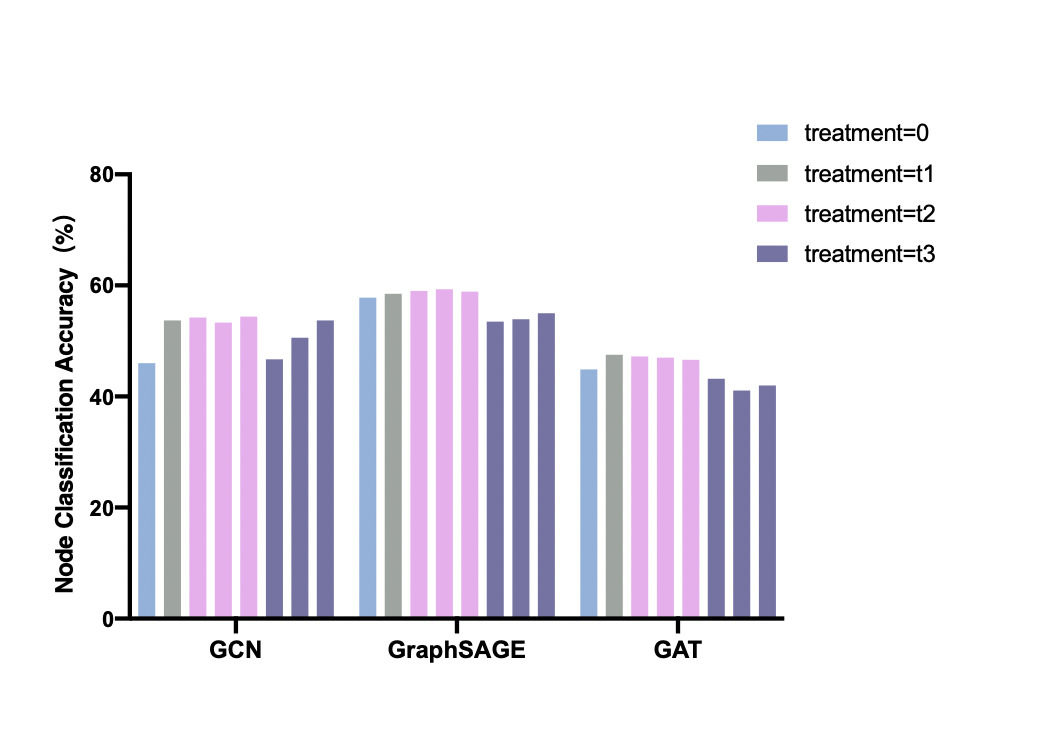}\label{fig6a}}\hspace{5pt}
	\subfloat[Squirrel]{\includegraphics[width=.45\columnwidth]{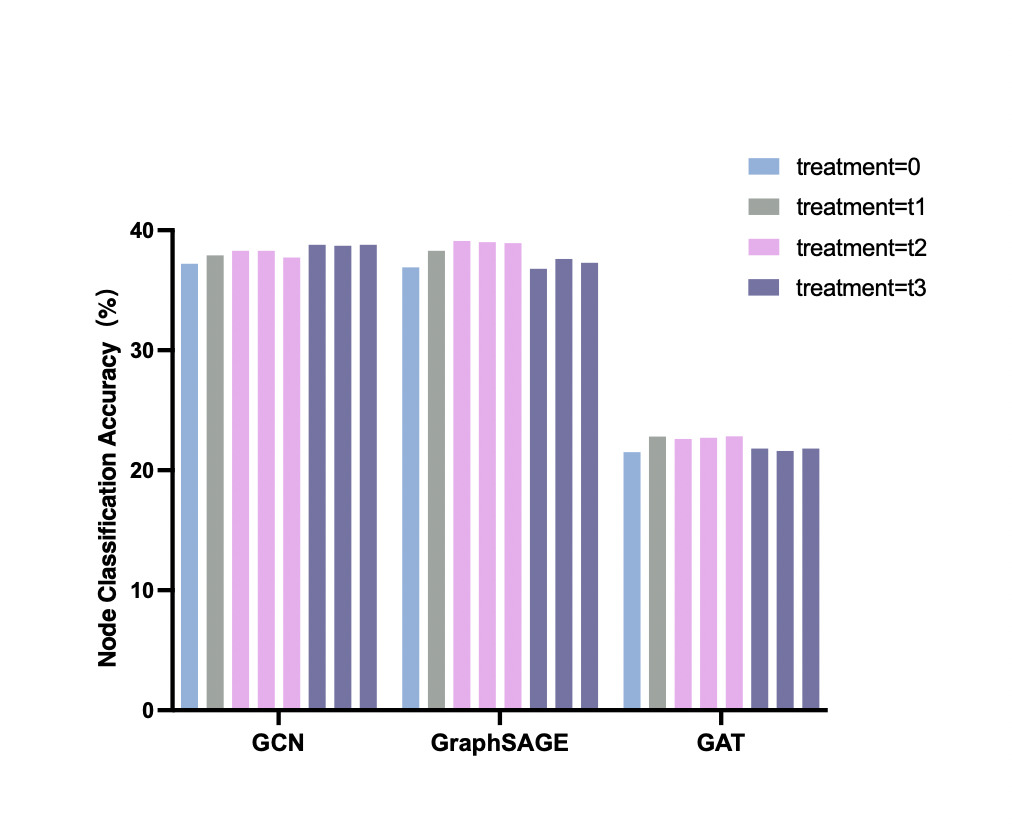}\label{fig6b}}
	\caption{The results of the control experiments.}
        \label{fig6}
\end{figure}

The results of the control experiments are illustrated in Figure \ref{fig6}. The following can be noted: 
\begin{enumerate}
    \item $Y\left(t_0\right) \approx Y\left(t_3\right)$. This indicates that the connections between different classes are substitutable, and both $W$ and $D$ are held constant. However, changing the nodes specifically connected in the LND has little effect on the discrimination ability of the GNN. It also indirectly indicates that nodes in the same class provide similar semantic information. We refer to this type of semantic information as \textbf{Class-Level Semantic}.
    \item $Y\left(t_1\right) \approx Y\left(t_2\right)>Y\left(t_0\right)$. After removing a portion of the neighbors while keeping $W$ constant, the GNN can better discriminate graph nodes; This improvement can be achieved by reducing $D$ and randomly altering $W$, which indicates that some connections in the graph are meaningless, and the distribution of such meaningless connecting edges does not vary with different classes.
\end{enumerate}

\subsection{Low Distraction and High Self-attention\label{sec4.2}}
\begin{figure}
	\centering
	\includegraphics[width=5in]{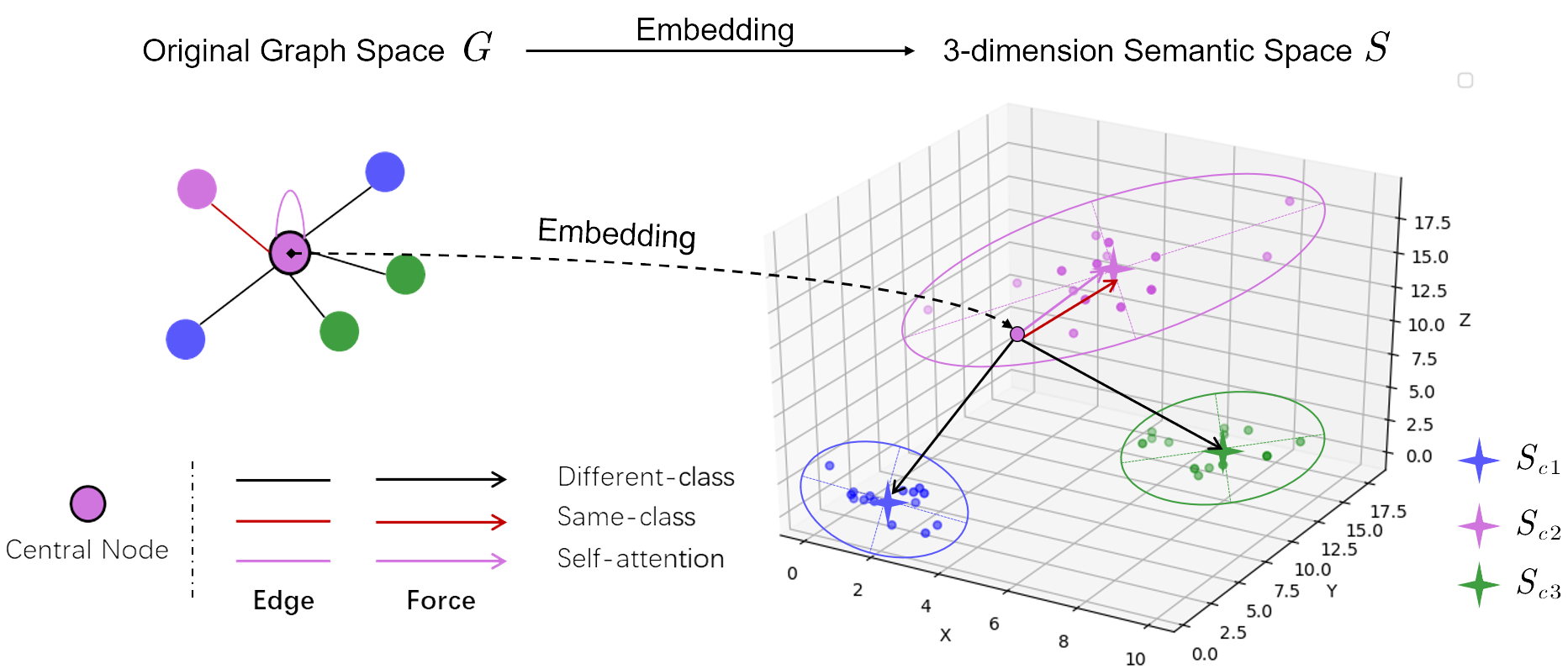}
	\caption{An example of an ideal semantic space (dimension = 3). Circles with different colors represent the distribution of the Class-level Semantic Cluster of different classes, and the quadrilateral star represents the center of each Class-level Semantic Cluster. The arrows in the representation space indicate the forces acting between nodes, whose strength is determined by the attention allocated by the central node.}
	\label{figadd1}
\end{figure}

Drawing on the observations from the pre-experiments, we propose a hypothesis and its inference. As illustrated in Figure \ref{figadd1}, for a highly heterophilic graph $G$ with nodes belonging to three classes, the ideal semantic space is composed of three compact clusters, and each cluster is composed of the mapped graph nodes belonging to their associated class. Each cluster has unique size, density, and location parameters, and the clusters can be easily distinguished from others. Observing more nodes of each class contributes to a more accurate distribution of its Semantic Clusters. Using a limit-thinking approach, if all nodes of a certain class are available, the Semantic Cluster observed can represent the distribution of all nodes belonging to that class. We refer to this as a Class-level Semantic Cluster, and the ideal space is referred to as Class-level Semantic Space.

\textbf{Hypothesis 1: Class-level Semantic Space Hypothesis.\label{Hyp1}} A graph $G$ can be mapped to an ideal $d$-dimensional semantic space $S=f(G)$, where belonging to the same class are located very close and nodes of different classes are as far away as possible. Since an ideal Semantic Cluster is compact, the cluster center can serve as the representation of nodes belonging to that class in the semantic space. Therefore, the connections between different classes are substitutable, where each Semantic Cluster center is denoted as Eq.\ref{equal5}:

\begin{equation}
\label{equal5}
\begin{aligned}
& s_c=\left\{\frac{\sum\limits_{v \in c} s_{v_1}}{n_c}, \frac{\sum\limits_{v \in c} s_{v_2}}{n_c}, \ldots \frac{\sum\limits_{v \in c} s_{v_d}}{n_c}\right\}, c \in C, \\
& \text { s.t. } \sum\limits_{v \in c}\left|s_v-s_c\right| \rightarrow 0 , \frac{1}{\left|s_c-s_j\right|} \rightarrow 0, j \in \complement_c C
\end{aligned}
\end{equation}

In the Class-level Semantic Space, it is evident that the closer the central node is to its Semantic Cluster center, the stronger its discrimination ability will be. Since the message passing mechanism aggregates features from neighboring nodes to the central node, neighbors from different classes exert a force that pushes the central node away from its own Semantic Cluster center, which is a distraction and should be minimized. Conversely, both neighbors from the same class and self-attention generate forces that pull the central node closer to its own Semantic Cluster center, which should be reinforced. In a highly heterophilic graph with few same-class neighbors, it is essential to enhance self-attention to mitigate distraction.

\textbf{Inference 1: Low Distraction and High Self-attention.\label{Inf1}} When a node in heterophilic graphs makes more use of its own information and ignores information derived from nodes of different classes during aggregation, its final representation will be closer to the Semantic Cluster center of its class in $S$.

\textbf{Proof.} For the central node $v_i$ and its neighboring node $v_j$, letting the aggregation weight be $w$, we can obtain the representation of $v_i$ after the aggregation and updating process as $h_i=\sigma\left(w_i \cdot z_i+\sum\limits_{j \in N_i} w_j \cdot z_j\right)$, where $h_i$ is closer to $s_{ci}$, model’s discrimination ability for $v_i$ will be stronger. When the graph is highly heterophilic, $h_i$ can be represented as Eq.\ref{equal6}: 

\begin{equation}
\label{equal6}
\begin{gathered}
w_i \cdot z_i+\sum\limits_{j \in N_i} w_j \cdot z_i=w_i \cdot z_i+\sum\limits_{m \in j, v_m \in c_i} w_m \cdot z_m+\sum\limits_{n \in j, v_n \notin c_i} w_n \cdot z_n \\
\rightarrow w_i \cdot s_c+\sum\limits_m w_m \cdot s_c+\sum\limits_n w_n \cdot z_n \\
\rightarrow\left(w_i+\sum_m w_m\right) \cdot s_c+\sum\limits_n w_n \cdot z_n \\
\text { s.t. } m \leqslant n ; w_i+\sum\limits_m w_m+\sum\limits_n w_n=1
\end{gathered}
\end{equation}

Because we hope $h_i$ is closer to $s_{ci}$, the optimization target is $\max \left(w_i+\sum\limits_m w_m-\sum\limits_n w_n\right)$. For the sake of the heterophilic graph condition stating that $m\leq n$, we hope that the weight of the central node itself $w_i$ can be maximized, which is equal to enhancing the self-attention level and avoiding the distraction caused by dissimilar neighbors.

GAT adaptively learns the weights of nodes to guide the aggregation. On the one hand, it may be easier to pose the distraction crisis to the central node due to the high proportion of interclass edges in a heterophilic graph. On the other hand, by learning a weight distribution with Low distraction and High Self-attention, GAT can directly enhance its discrimination ability. Therefore, we foster strengths and circumvent weaknesses for GAT by leveraging the learned attention distribution as signals, to guide the GAT to identify and remove the Distraction Neighbors. The graph trimming operation does not require architecture alternations or new neighbor searches but rather learns an optimal attention distribution to enhance self-attention.

\section{Methodology\label{sec5}}
\subsection{The Architecture of CAT}
\begin{figure}
	\centering
	\includegraphics[width=6in]{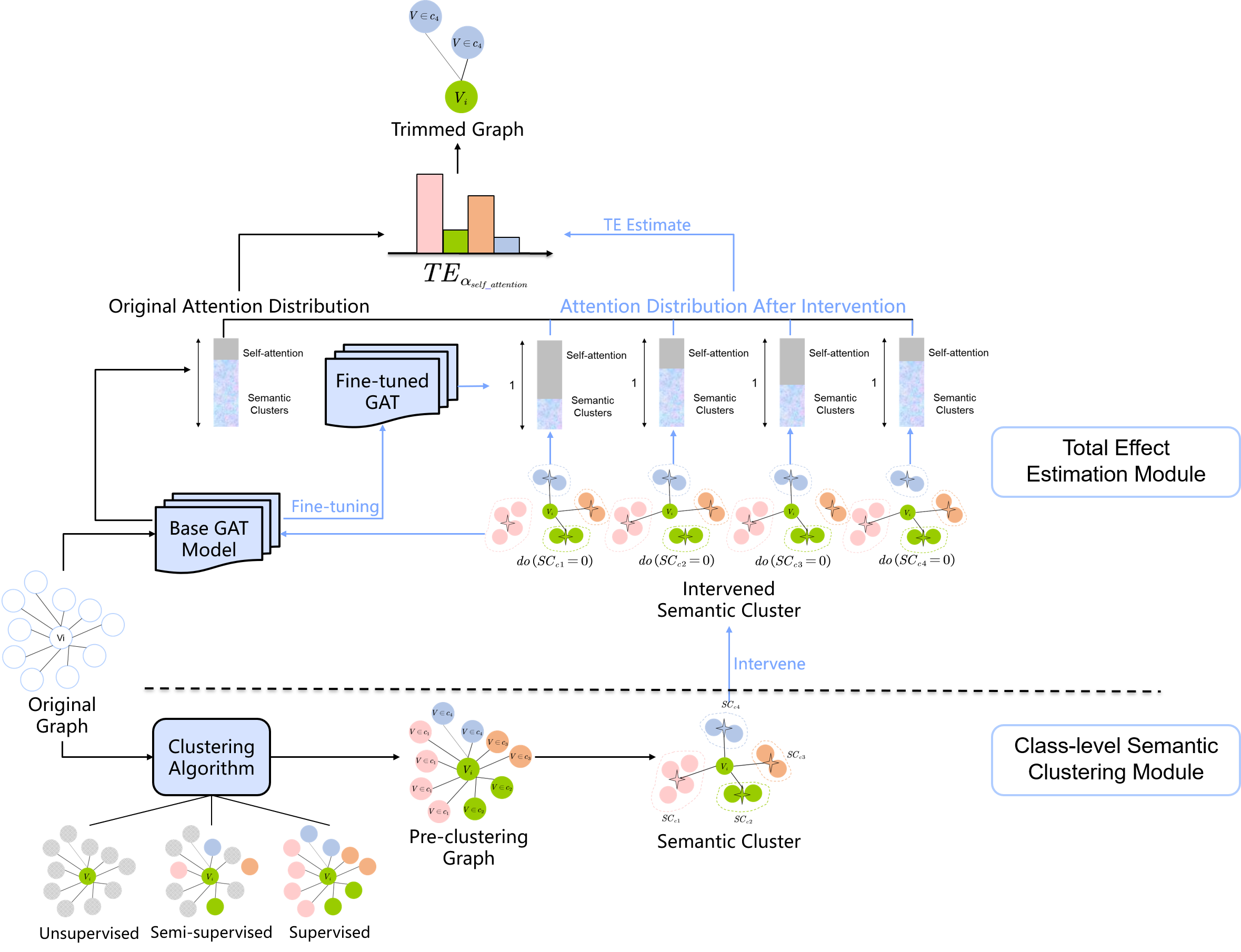}
	\caption{The pipeline of CAT.}
	\label{fig7}
\end{figure}

The Causal graph Attention network for Trimming heterophilic graphs (CAT) proposed in this paper mainly contains two important modules: the Class-level Semantic Clustering Module and the Total Effect Estimation Module. The former obtains the basic unit for estimating the TE of the neighboring nodes, and the latter further estimates the TE via graph intervention. We introduced the CAT in Algorthim \ref{algorithm1}, where the $\Theta_W$ and $\Theta_{W2}$ of the GAT represent the model parameters for feature transformation and attention distribution learning, respectively. The pipeline of CAT is illustrated in Figure \ref{fig7}. As illustrated in Figure \ref{fig7}, the framework of CAT can adopt different GATs as the base model, and finally obtain a trimmed graph that can optimize the attention distribution of the base GAT. 

\begin{enumerate}
    \item \textbf{Class-level Semantic Clustering Module.} This module is derived from the \hyperref[Hyp1]{Class-level Semantic Space Hypothesis}, which maps the LND of the central node to a space that can better discriminate class-level semantics. The Semantic Clusters output in this module further serve as the basic object for estimating TE.
    \item \textbf{Total Effect Estimation Module.} This module is derived from the \hyperref[Inf1]{Low distraction and High Self-attention}, which obtains the TE of each class on the central node by intervening in different Semantic Clusters. The Distraction Neighbors are identified in accordance with the TE and removed to obtain the final trimmed graph.
\end{enumerate}

\IncMargin{1em}
\begin{algorithm}
  \SetKwData{Left}{left}\SetKwData{This}{this}\SetKwData{Up}{up}
  \SetKwFunction{Union}{Union}\SetKwFunction{FindCompress}{FindCompress}
  \SetKwInOut{Input}{input}\SetKwInOut{Output}{output}
  \Input{$G=(V,E)$, $A \in \mathbb{R}^{N \times N}$, $X \in \mathbb{R}^{N \times F}$, $C$, $epoch_{pretrain}$, $epoch_{finetuning}$, $f_{clustering}$, initialized $\Theta_W, \Theta_{W2}$ of $f_{GAT}$}
  \Output{$A_{Trim} \in \mathbb{R}^{N \times N}$}
  \BlankLine
  \tcp*[h]{Class-level Semantic Clustering}\;
    $SC=f_{\text{clustering}}(X)$, $SC \in \mathbb{R}^{n \times 1}$\;
  \tcp*[h]{Pretrain the base GAT}\;
    \For{$epoch$ in $epoch_{pretrain}$}
        {$Z, \alpha_{SC}, \alpha_{self\_attention}=f_{\text{GAT}}(A,X)$\;
        Update $\Theta_W, \Theta_{W2}$ of $f_{GAT}$\;}
    Freeze $\Theta_W$ and re-initialize $\Theta_{W2}$\;
  \tcp*[h]{Semantic Cluster intervention}\;
    \For{$c$ \text{in} $C$}{
        \For{$V_i$ \text{in} $V$}{
            \If{$A_{ij}=1$ and SC(j)=c}{
                $\hat{A}_{i j}^c=0$ \;
            }
        }
    \tcp*[h]{Intervened attention learning}\;
    \For{$epoch$ in $epoch_{finetuning}$}{
        $Z^c, \alpha_{SC}^c, \alpha_{self\_attention}^c=f_{\text{GAT}}(\hat{A}^c,X)$\;
        Update $\Theta_{W2} \longrightarrow f_{GAT}^c$\;
    }
    }
  \tcp*[h]{Graph trimming}\;
      \For{$c$ \text{in} $C$}
      {$TE_{\alpha_{self\_attention}}^c=\alpha_{self\_attention}^c-\alpha_{self\_attention}$\;
      }
      \For{$V_i$ \text{in} $V$}{
            \If{$A_{ij}=1$ and $SC(j)=\min \left(TE_{\alpha_{self\_attention}} \right)$}{
                $A_{ij}^{Trim}=1$ \;}
            }
  \caption{Causal graph Attention network for Trimming heterophilic graphs(CAT)}\label{algorithm1}
\end{algorithm}\DecMargin{1em}

\subsection{Class-level Semantic Clustering Module}\label{module1}
Based on the \hyperref[Hyp1]{Class-level Semantic Space Hypothesis}, we consider that the neighbors impact the self-attention learning of the central node with their classes as the basic units. This idea is very intuitive for heterophilic graphs; when the representations of graph nodes are difficult to distinguish, observing more nodes for a class makes it easier to obtain the global distribution of that class.

\begin{figure}
	\centering
	\includegraphics[width=4in]{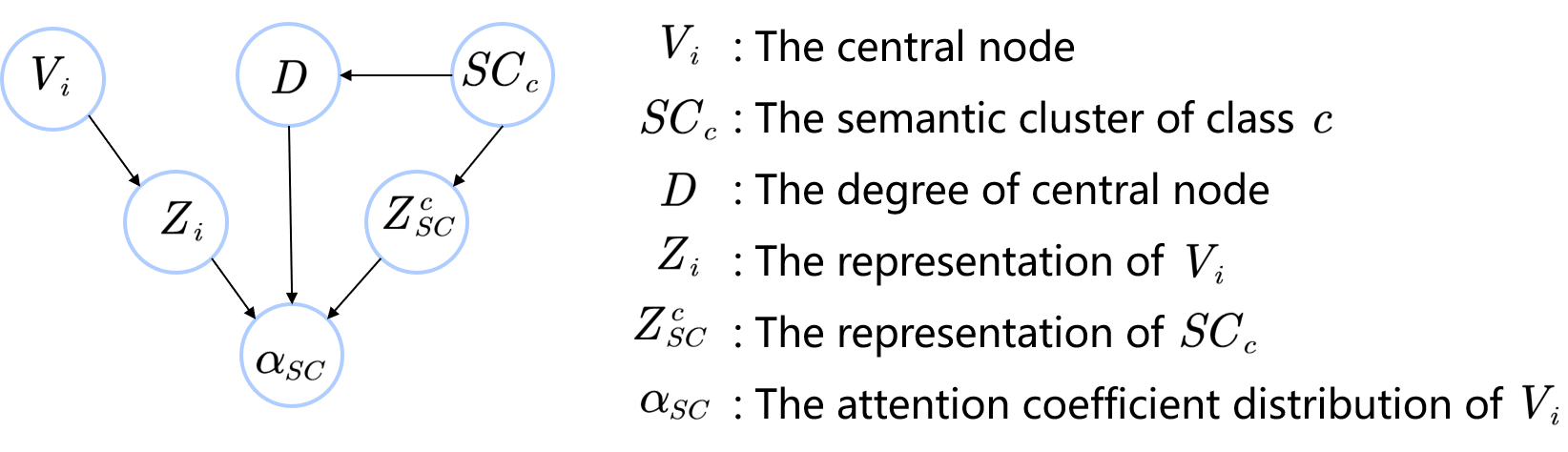}
	\caption{Causal graph behind the GAT at the Semantic Cluster level.}
	\label{fig8}
\end{figure}

In that semantic space, we treat the local neighbors belonging to the same cluster as a whole, which is referred to as Semantic Cluster $SC=f_{\text {clustering }}(x), \quad x \in \mathbb{R}^{n \times F},  \quad SC \in \mathbb{R}^{n \times 1}$. Where $SC(i) \in C$ represents the cluster class of nodes with the index $i$. Accordingly, the center of each SC in that semantic space is $SC_c=\left\{\frac{\sum\limits_{SC(v)=c} SC_{v_1}}{n_c}, \frac{\sum\limits_{SC(v)=c} SC_{v_2}}{n_c}, \ldots \frac{\sum\limits_{SC(v)=c} SC_{v_d}}{n_c}\right\}, c \in C$. We can update the causal graph proposed in Figure \ref{fig2} to Figure \ref{fig8}. 

\begin{figure}
	\centering
	\includegraphics[width=3in]{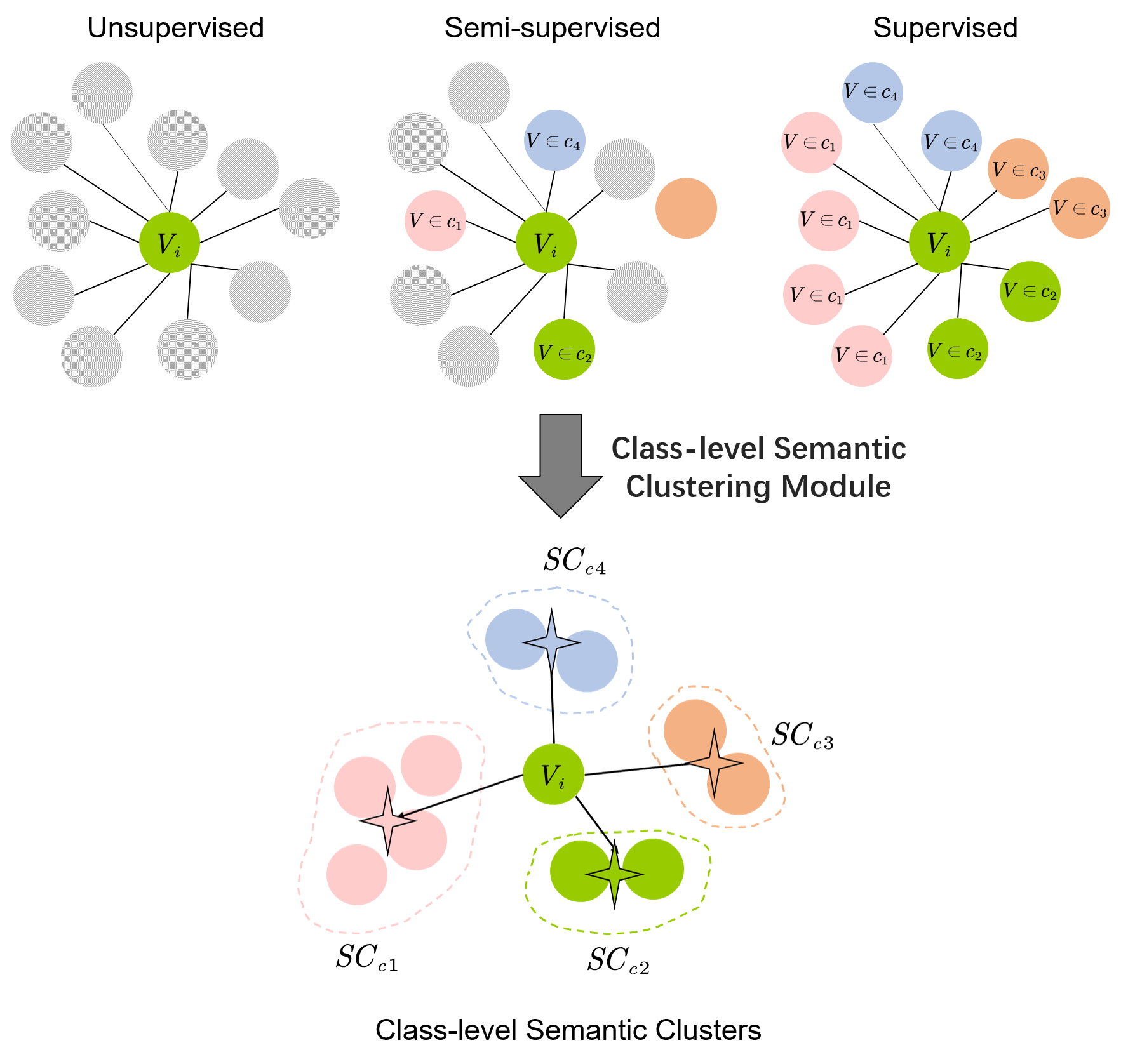}
	\caption{Three paradigms adopted in the Class-level Semantic Clustering Module.}
	\label{figadd2}
\end{figure}

As shown in Figure \ref{figadd2}, three learning paradigms can be adopted in this module to obtain Class-level Semantic Clusters. Ordered in ascending prior knowledge about node category distribution they are: unsupervised, semi-supervised, and supervised learning. The more information we acquire about the categorical distribution of graph nodes, the closer the obtained Semantic Cluster distribution will be to the distribution in the ideal semantic space. This further indicates a more accurate estimate of the total effect of attention-learning on Class-level Semantic Clusters corresponding to each category. We construct three CAT variants by adopting the following three learning paradigms in this module:

\begin{itemize}
    \item \textbf{Unsupervised manner:} For all nodes in the graph, their categorical labels are unseen. Unsupervised clustering methods can be employed to obtain a rough semantic space with the input of node features. The CAT variant built in this manner is referred to as \textbf{CAT-unsup}.
    \item \textbf{Semi-supervised manner:} Categorical labels for a fixed ratio of nodes are known and used to infer the labels of unknown nodes. Classification methods with a semi-supervised setting can be employed to obtain a less rough semantic space with the input node features. The CAT variant constructed in this manner is referred to as \textbf{CAT-semi}.
    \item \textbf{Supervised manner:} Categorical labels for all nodes are known and their categorical distribution is completely and accurately observed. It should be noted that the label information is only available in the Class-level Semantic Clustering stage and is not used for node classification. The CAT variant employed in this manner is referred to as \textbf{CAT-sup}.
\end{itemize}

\subsection{Total Effect Estimation Module}\label{module2}
As illustrated in Figure \ref{fig8}, there are two paths from class-level Semantic Clusters to the central node’s attention distribution that will jointly influence the representation learning of the central node. Therefore, we employed \hyperref[TE]{total effect} as a measurement of the Distraction Effect based on the preliminary of \hyperref[CI]{Causal Inference}. This module contains three important steps, semantic cluster intervention, intervened attention learning, and graph trimming.

\begin{figure}
	\centering
	\includegraphics[width=3in]{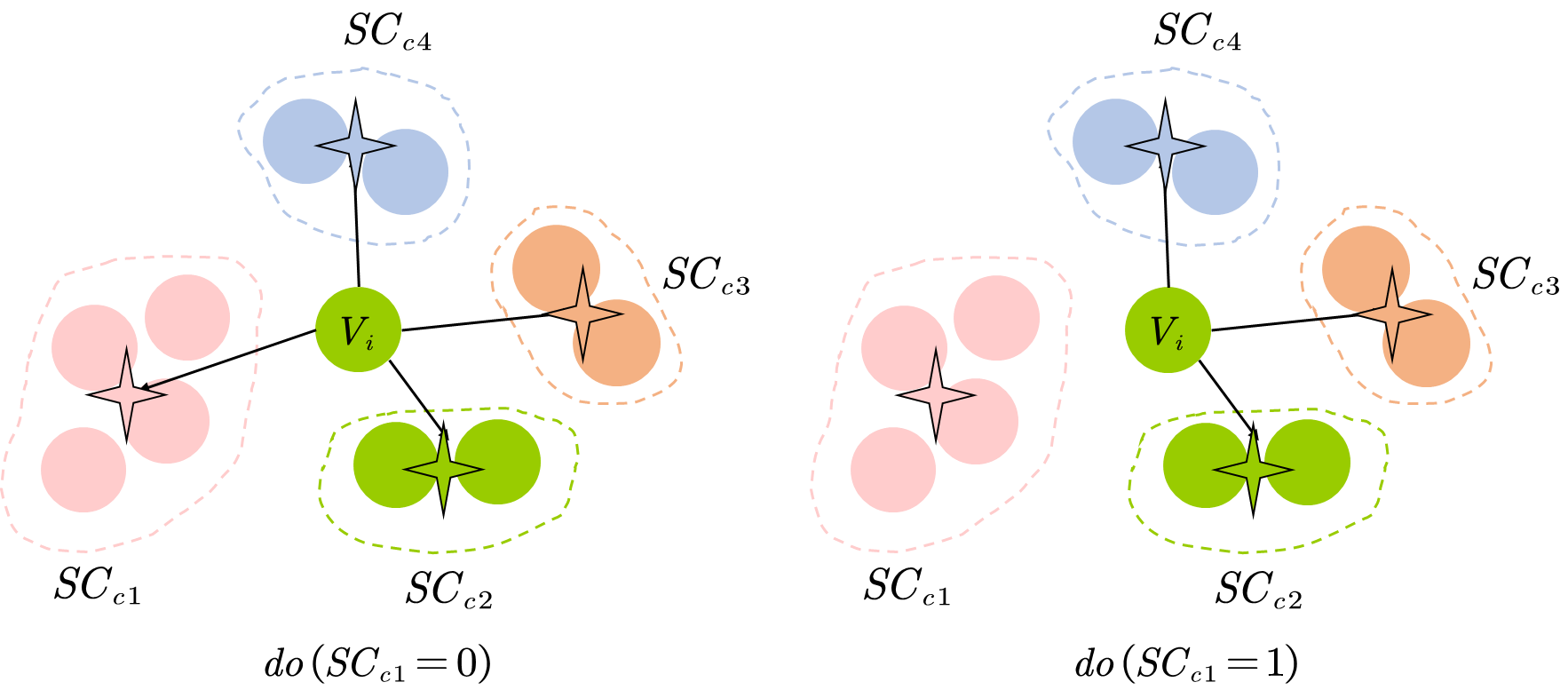}
	\caption{Semantic cluster intervention.}
	\label{fig9}
\end{figure}

\begin{enumerate}
    \item \textbf{Semantic Cluster Intervention.} As detailed in Section \ref{sec3}, Total effect is estimated based on the intervention.This step is theoretically equivalent to forcing the central node to answer a causal question: \textbf{how will my attention distribution change if Semantic Cluster $c$ is removed from my LND?} The physical intuition behind this intervention-related question is that it is an operation that renders the nodes belonging to Semantic Cluster $c$ invisible to the central node. Figure \ref{fig9} can be mathematically modelled as Eq.\ref{equal7}, where $A$ represents the adjacency matrix of the original graph, and $\hat{A}$ represents that of the intervention graph.
        \begin{equation}
        \label{equal7}
     \begin{aligned}
        & do\left(SC_c=0\right): A_{ij}=1, SC(j)=c \\
        & do\left(SC_c=1\right): \hat{A}_{ij}^c=0, SC(j)=c
    \end{aligned}
        \end{equation}
        
    \item \textbf{Intervened Attention Learning.}  Since the intervention in the Semantic Cluster does not affect the shape of Class-level Semantic Space (which is governed by the data generation mechanism of the graph and considered to be invariant), we need to guarantee that before and after the intervention, the base GAT only changes its attention distribution, while the other capabilities remain unchanged. From the model implementation perspective, we do not alter the parameters responsible for transforming node features, allowing the model to solely reallocate the attention assigned to neighboring nodes and itself. The attention assigned to Semantic Clusters can be represented as Eq.\ref{equal8}, where $\alpha_{SC}^c=\sum\limits_{SC(V_j)=c} \alpha_{ij}$, and the self-attention of central node $V_i$ is $\alpha_{self\_attention }=1-\sum\limits_C \alpha_{SC}^c$.
        \begin{equation}
        \label{equal8}
     \alpha_{SC}=\left\{\alpha_{SC}^1, \alpha_{SC}^2, \ldots, \alpha_{SC}^C\right\} \in \mathbb{R}^{C \times 1}
        \end{equation}
    
    \item \textbf{Graph Trimming.} According to the concept in Section \ref{sec3}, we can calculate the TE of Semantic Cluster $c$ based on the self-attention of the central node according to Eq.\ref{equal9}.
            
        \begin{equation}
        \label{equal9}
        \begin{split}
    TE_{\alpha_{self\_attention}} & = E_{\alpha_{self\_attention} \mid do\left(SC_c=1\right)}\left[\alpha_{self\_attention } \mid \operatorname{do}\left(SC_c=1\right)\right] \\
     & - E_{\alpha_{self\_attention} \mid d o\left(S C_c=0\right)}\left[\alpha_{self\_attention } \mid \operatorname{do}\left(SC_c=0\right)\right] 
        \end{split}
        \end{equation}
                
    The lower the value of the sum of $TE_{\alpha_{SC}}$ is, i.e., the higher the value of $TE_{\alpha_{self\_attention}}$ is, the more it can distract the central node and lead to low self-attention for the central node, and vice versa. Therefore, we remove the Semantic Cluster with lower $TE_{\alpha_{SC}}$ values and retain only the Semantic Cluster with the highest $TE_{\alpha_{SC}}$. In other words, only the Semantic Cluster with the lowest TE on self-attention of the central node will remain. Eventually, we obtain the adjacency matrix of the trimmed graph denoted as Eq.\ref{equal10}, which equals an operation that removes the edges connecting Distraction Neighbors and the central nodes. 
        \begin{equation}
        \label{equal10}
     A_{\text {Trim}}=\left\{a_{ij}=1, SC(j)=\min \left(TE_{\alpha_{self\_attention} }\right)\right\}
        \end{equation}
\end{enumerate}

\section{Experiments and Results\label{sec6}}
\subsection{Databases \label{sec6.1}}
To ensure the richness and representativeness of the employed data, we selected seven heterophilic graphs of three sizes. All datasets possess an Edge Homophily below 0.23. The basic information of the datasets is given in the Table \ref{tab2}. Edge Homophily is defined in Eq.\ref{equal11}.

\begin{equation}
\label{equal11}
H_{\text {edge }}=\frac{\left|y_i=y_j,(i, j) \in E\right|}{|E|}
\end{equation}

\begin{itemize}
    \item \textbf{Small-size datasets.} We use the WebKB dataset \cite{37} constructed from the WebKB web page. It was collected from the computer science departments of Cornell, Texas, and Wisconsin-Madison University. This dataset was built from the hyperlinks between web pages, and the features of nodes are the bag-of-words representations. The nodes belong to five categories.
    \item \textbf{Medium-size datasets.} We use the Chameleon and Squirrel datasets \cite{37} collected from Wikipedia, which are applicable for node regression and node classification tasks. In these datasets, the nodes represent web pages and the edges are links between them. When applying them for the node classification tasks, the target is to predict 5 classes based on the average traffic of web pages.
    \item \textbf{Large-size datasets.} We use the Roman-empire \cite{55} dataset constructed from Wikipedia articles. In this dataset, the nodes represent words in a text, and the edges are constructed from their context. The target is to predict 18 classes based on the syntactic role of the nodes.
\end{itemize}

\begin{table}
    \caption{Statistics of the datasets.}
    \label{tab2}
    \centering
        \begin{tabular}{cccccccc}
        \toprule
        \multicolumn{2}{c}{Dataset}  & Nodes  &Edges  &Average Degree  &Features  &Classes  &Edge Homophily  \\
        \midrule
        \multirow{3}{*}{Small-size}  &Cornell  &183	&295	&3.06	&1703	&5	&0.13 \\
            &Texas  &183	&309	&3.22	&1703	&5	&0.11  \\
            &Wisconsin  &251	&499	&3.71	&1703	&5	&0.20 \\
        \midrule
        \multirow{3}{*}{Medium-size} &Chameleon  &2277	&62792	&27.60	&128	&5	&0.23 \\
                    &Squirrel &5201	&396846	&78.33	&128	&5	&0.22\\
                    &Actor  &7600	&33269	&7.02	&932	&5	&0.21\\
        \midrule
        \multirow{1}{*}{Large-size}  &Roman-empire  &22662  &65854  &2.91  &300  &18  &0.04  \\
        \bottomrule
        \end{tabular}
\end{table}

\subsection{Experiments\label{sec6.2}}
We aim to explore the effect of neighboring nodes on the central node's attention-learning within the GAT mechanism. Therefore, we take a GAT with fixed architecture (which can be regarded as possessing the same aggregation and feature transformation abilities) as the base model and compare its discrimination ability on graphs with different LNDs. We focus more on the difference caused by attention distribution learned by the GAT, thus choosing not to tune the architecture of the GAT through careful optimization and parameter tuning. To better demonstrate the effectiveness of CAT, we examined its efficacy across different GATs and adopted three GATs as base models. They are built with distinctive motivations and improved mechanisms and thus can present different scenarios. To distinguish between the results of different base models, we replace "G" in the original name of base models with "C" to represent their corresponding CATs. The base GATs and their fixed architecture are as follows:

\begin{itemize}
    \item \textbf{GAT} \cite{20}. The originally proposed Graph Attention Network implicitly specifies different weights for neighboring nodes. It injects a graph structure into its self-attention strategy to learn attention coefficients, thus learning node representations in a more informative way. We set the number of GAT layers to 2 and the number of heads to 8.
    \item \textbf{GATv2} \cite{gatv2}. A dynamic graph attention variant that can learn dynamic attention by simply switching the order of internal operations in GAT. It can outperform the original GAT when more complex interactions are observed between nodes in the input graph. We set the layer of GATv2 to 2 and the number of heads to 8.
    \item \textbf{GATv3} \cite{42}. A new attention mechanism that calculates the query and key from other GNN models. It can be adaptively used with homophilic or heterophilic graphs. We set the GATv3 layer to 2 and adopt a one-layer GCN in the K and Q modules. To better investigate the effect of attention-learning, we fix the weight of the calculated attention to 1 and abandon the original weighted attention strategy. 
\end{itemize}

For all base GATs and their CAT variants, we use the Adam optimizer with a learning rate of 0.001 and a weight decay of 0.0001 to train the model. A single Nvidia 2080Ti GPU was used for training with a negative log likelihood loss. The maximum number of iterations was 600, and the tolerance of the early stopping strategy based on the classification accuracy on the validation set is set to 50. To evaluate the model accuracy, we divided each dataset into training, validation, and test sets at a ratio of 6:2:2 and used the average classification accuracy and standard deviation attained on the test set over 100 repetitions as the final evaluation metrics. We set the dimensions of the hidden layers to \{16,32,64,128\} and adopted the optimal classification accuracy. We conduct comparison and ablation experiments to verify the validity of the architecture and individual modules of CAT, respectively. Visualization experiments were also carried out to further interpret the results.

\subsubsection{Comparison Experiment\label{sec6.2.1}}
We feed the original heterophilic graph and the trimmed graph obtained by CAT variants to the base GAT to obtain the final node classification accuracy. The trimmed graphs were obtained by using three variants of CAT with the following settings:

\begin{itemize}
    \item \textbf{CAT-unsup}. Since the number of Semantic Clusters is known (equal to the number of target classes), we use the K-means++ algorithm in the Class-level Semantic Clustering Module in an unsupervised manner. To avoid the influence of the initial clustering centers on the results, we used 0,10,100 as random seeds for the initial clustering centers in K-means++.
    \item \textbf{CAT-semi}. For the semi-supervised manner, we employed a two-layer Multi-Layer Perception (MLP) to learn the categorical distribution of nodes. To maintain the consistency of the semi-supervised node classification task, we use the same dataset split described in Section \ref{sec6.1} for the MLP.
    \item \textbf{CAT-sup}. In a supervised manner, we directly used the labels to generate the Class-level Semantic Clusters.
\end{itemize}


The results are shown in Table \ref{tab3}. Our approach exhibits improvements across all base GAT models. Even on the large-size dataset with an Edge Homophily level of only 0.04, the minimum relative improvements for GAT and GATv2 are 13.5\% and 10.1\%, respectively. Adopting semi-supervised and fully-supervised paradigms can lead to further improvements. 

Additionally, we observed performance shifts on different base GATs. There are slight differences between the performances of GAT and GATv2 across most datasets. However, on the Roman-empire dataset, GATv2 outperforms GAT by over two percentage points. The potential reason could be that Roman-empire dataset contains more node categories and a larger graph size, resulting in more complex interactions between nodes in the graph, at which point the dynamic attention captured by GATv2 proves effective.

GATv3 exhibits the best performance among all base models due to its incorporation of a new attention mechanism that leverages other GNN models, thereby enhancing its discrimination capability. However, CAT can further improve its classification accuracy. Among all base models, the relative improvement provided by CAT for GATv3 is the lowest. The reason is that GATv3 already boasts comparatively high discrimination capabilities on heterophilic graphs, making further enhancement more challenging. Each of the three base models exhibits strengths in different scenarios, yet CAT demonstrates the capability to further boost their performance across all datasets.

In terms of the standard deviation of the prediction accuracy, on small-size datasets, the deviation of CATs is relatively large compared with that of the base GATs. However, on medium-size and large-size datasets, CAT significantly reduces the deviation and achieves more stable and statistically significant predictions.

For all base GATs, CAT-sup generally outperforms CAT-unsup and CAT-semi. This is because it leverages more information in the Class-level Semantic Clustering Module, thereby obtaining a more accurate distribution of Semantic Clusters. This speculation can also explain why CAT-unsup performs worst and the CAT-semi consistently performs at a moderate level. On the one hand, this indicates that precise Class-level Semantic Clustering can facilitate better attention allocations. On the other hand, it underscores the challenge of learning better Semantic Spaces. CAT-unsup variants with different random seeds also achieve significantly different performances. For example, although all CAT-unsup models can outperform the GAT on the Wisconsin dataset, CAT-unsup with a random seed value of 10 attains an accuracy that is over 10\% lower than that produced with a value of 100. CATv2-unsup exhibits a similar pattern on the Texas dataset. This indicates that we can barely guarantee that the learned Class-level Semantic Space is optimal or is approaching optimal for unsupervised learning purposes. Additionally, the results indicate that the output of Class-level Semantic Clustering plays a significant role in the overall method.

\begin{table}[]
\caption{Node classification accuracy. 0, 10, and 100 represent the corresponding random seeds used in the unsupervised clustering method. The best and worst results achieved by the CAT variants are marked in \textbf{bold} and \uwave{wavy line}, respectively, and their relative improvements over GAT are shown below. The base models and the worst result among all models are marked in red. OOM represents out-of-memory.}
\label{tab3}
    \centering
\tabcolsep=0.1cm
\begin{tabular}{ccccccccc}\\
\toprule
\multicolumn{2}{c}{\multirow{2}{*}{\textbf{Dataset}}}    & \multicolumn{3}{c}{\textbf{Small-size}} & \multicolumn{3}{c}{\textbf{Medium-size}} & \textbf{Large-size}   \\
\multicolumn{2}{c}{($H_{edge}$)}         &\makecell{Cornell \\(0.13)}  & \makecell{Texas \\(0.11)}   & \makecell{Wisconsin \\(0.20)}  & \makecell{Chameleon \\(0.23)}  & \makecell{Squirrel \\(0.22)}  & \makecell{Actor \\(0.21)}  & \makecell{Roman-empire \\(0.04)}\\
\midrule
\multicolumn{2}{l}{\textcolor[rgb]{1,0,0}{GAT}}                       &\textcolor[rgb]{1,0,0}{60.9$\pm$3.4}          &\textcolor[rgb]{1,0,0}{49.9$\pm$2.2}        &\textcolor[rgb]{1,0,0}{53.7$\pm$2.5}            &\textcolor[rgb]{1,0,0}{44.9$\pm$1.8}            &\textcolor[rgb]{1,0,0}{21.5$\pm$1.7}           &\textcolor[rgb]{1,0,0}{28.6$\pm$0.4}        &\textcolor[rgb]{1,0,0}{54.2$\pm$0.5}              \\
\multicolumn{1}{l}{\multirow{3}{*}{CAT-unsup}}       & 0         &75.8$\pm$3.5          &\uwave{65.2$\pm$2.4}        &\uwave{62.5$\pm$4.3}            &\uwave{48.8$\pm$0.6}            &29.0$\pm$0.3           &32.6$\pm$0.6        &\uwave{61.5$\pm$0.2}               \\
 & 10        &72.9$\pm$4.1          &70.2$\pm$2.2        &70.2$\pm$2.2            &51.9$\pm$0.7             &28.9$\pm$0.3           &33.7$\pm$0.6        &63.5$\pm$0.3               \\
 & 100       & \uwave{69.0$\pm$2.0}         &69.6$\pm$3.4        &69.6$\pm$3.4            &51.9$\pm$1.0            &\uwave{28.4$\pm$0.3}           &\uwave{31.5$\pm$0.4}        &62.2$\pm$0.2              \\
\multicolumn{2}{l}{CAT-semi}                  &71.0$\pm$3.2          &73.0$\pm$3.9        &73.0$\pm$3.9            &50.6$\pm$0.5            &28.7$\pm$0.4           &32.8$\pm$0.6        &61.9$\pm$0.2              \\
\multicolumn{2}{l}{CAT-sup}                   & \textbf{80.4$\pm$3.0}        &\textbf{76.7$\pm$3.1}        &\textbf{82.0$\pm$1.6}            &\textbf{53.4$\pm$0.9}            &\textbf{32.4$\pm$0.9}           &\textbf{35.5$\pm$0.5}        &\textbf{64.4$\pm$0.2}              \\
\midrule
\multicolumn{2}{l}{Relative Improvement (\%)}      & 13.3-32.0         &30.9-53.7        &16.4-52.7            &8.7-18.9            &32.1-50.7           &10.1-24.1        &13.5-18.8              \\
\midrule
\multicolumn{2}{l}{\textcolor[rgb]{1,0,0}{GATv2}}                       &\textcolor[rgb]{1,0,0}{61.1$\pm$3.6}          &\textcolor[rgb]{1,0,0}{50.2$\pm$2.2}        &\textcolor[rgb]{1,0,0}{53.8$\pm$2.4}            &\textcolor[rgb]{1,0,0}{45.9$\pm$1.6}            &\textcolor[rgb]{1,0,0}{21.4$\pm$2.1}           &\textcolor[rgb]{1,0,0}{28.5$\pm$0.4}        &\textcolor[rgb]{1,0,0}{56.5$\pm$0.8}              \\
\multicolumn{1}{l}{\multirow{3}{*}{CATv2-unsup}}      & 0        &78.1$\pm$3.2         &\uwave{62.8$\pm$2.9}        &77.3$\pm$1.5            &51.8$\pm$0.9           &28.2$\pm$0.4           &\uwave{31.8$\pm$0.5}        &63.3$\pm$0.1               \\
 & 10        &\uwave{74.5$\pm$4.3}          &75.8$\pm$1.5        &79.1$\pm$2.3            &52.0$\pm$0.7            &\uwave{28.0$\pm$0.4}           &32.4$\pm$0.5        &\uwave{62.2$\pm$0.2}              \\
& 100       &74.8$\pm$1.6        &70.4$\pm$4.9        &\uwave{76.9$\pm$3.0}            &\uwave{50.6$\pm$0.5}            &28.5$\pm$0.3           &32.3$\pm$0.5        &63.1$\pm$0.2              \\
\multicolumn{2}{l}{CATv2-semi}                  &81.5$\pm$3.4          &\textbf{75.3$\pm$3.4}        &78.7$\pm$2.2            &53.1$\pm$0.9            &29.9$\pm$1.4           &31.9$\pm$0.5        &63.0$\pm$0.2              \\
\multicolumn{2}{l}{CATv2-sup}                   &\textbf{81.7$\pm$3.8}       &72.8$\pm$2.0        &\textbf{84.2$\pm$2.0}            &\textbf{56.9$\pm$0.9}            &\textbf{32.4$\pm$1.3}           &\textbf{33.1$\pm$0.5}        &\textbf{63.4$\pm$0.2}              \\
\midrule
\multicolumn{2}{l}{Relative Improvement (\%)}      & 21.9-33.7         & 25.1-50.0       & 42.9-56.5           & 10.2-24.0          & 30.8-51.4          & 11.6-16.1       & 10.1-12.2             \\
\midrule
\multicolumn{2}{l}{\textcolor[rgb]{1,0,0}{GATv3}}                       &\textcolor[rgb]{1,0,0}{86.3$\pm$2.2}          &\textcolor[rgb]{1,0,0}{81.6$\pm$2.4}        &\textcolor[rgb]{1,0,0}{80.8$\pm$2.3}            &\textcolor[rgb]{1,0,0}{62.9$\pm$1.0}            &\textcolor[rgb]{1,0,0}{33.7$\pm$0.7}           &\textcolor[rgb]{1,0,0}{35.1$\pm$0.5}        & OOM             \\
\multicolumn{1}{l}{\multirow{3}{*}{CATv3-unsup}}     & 0        &88.2$\pm$2.0         &83.1$\pm$2.9        &\uwave{82.3$\pm$2.5}            &64.2$\pm$0.8            &53.7$\pm$0.9           &37.8$\pm$0.6        & -             \\
 & 10        &\uwave{87.5$\pm$2.0}          &\uwave{82.8$\pm$2.5}        &84.3$\pm$2.5            &64.2$\pm$0.9            &\uwave{53.6$\pm$0.8}           &\uwave{36.9$\pm$0.5 }       & -             \\
& 100       &88.0$\pm$2.2        &\textbf{83.3$\pm$3.4}        &83.2$\pm$2.4            &\uwave{63.4$\pm$0.8}            &\uwave{53.6$\pm$0.7}           &38.0$\pm$0.5        & -             \\
\multicolumn{2}{l}{CATv3-semi}                  &88.4$\pm$2.1          &\uwave{82.8$\pm$2.7}        &84.6$\pm$2.2            &67.1$\pm$0.8            &55.9$\pm$0.8           &37.7$\pm$0.6        &  -            \\
\multicolumn{2}{l}{CATv3-sup}                   &\textbf{88.8$\pm$2.1}       &83.0$\pm$2.5        &\textbf{85.6$\pm$2.1}            &\textbf{69.9$\pm$1.0}            &\textbf{59.3$\pm$1.8}           &\textbf{38.5$\pm$1.2}        &  -            \\
\midrule
\multicolumn{2}{l}{Relative Improvement (\%)}      &1.4-2.9         & 1.5-2.1        &  1.9-5.9          &  0.8-11.1          & 59.1-76.0          & 5.1-9.7       & -             \\
\bottomrule
\end{tabular}
\end{table}

\subsubsection{Ablation Experiment\label{sec6.2.2}}
To investigate the effectiveness of each component in the proposed method,  we conducted ablation studies on its two modules and accordingly obtained two trimmed graphs. For the sake of making a convincing comparison, we select CAT-unsup in this section because it performs the worst among the three variants of CAT. The results of the ablation experiment are shown in Table \ref{tab4}. 

\begin{enumerate}
    \item \textbf{CAT (random\_cluster).} To investigate the effectiveness of the Class-level Semantic Cluster module, we replace it with a randomly assigned cluster module. We set the random seeds to 0, 10, and 100 for the random cluster assignment. 
    \item \textbf{CAT(high\_distraction).} To investigate the effectiveness of the Total Effect Estimation, we remove the neighbors with lower distraction and create a High Distraction and Low Self-attention scenario. This model reserves the $\max \left(TE_{\alpha_{self\_attention}} \right)$ from the Total Effect Estimation.
\end{enumerate}

\begin{table}
\caption{The results of the ablation experiments. The CAT model in this table represents CAT-unsup. The best accuracy is marked in \textbf{bold}, and the worst is indicated with a \uwave{wavy line}.}
\label{tab4}
    \centering
\begin{tabular}{ccccc}
\toprule
\multirow{2}{*}{Dataset} & \multirow{2}{*}{Trimmed graph} & \multicolumn{3}{c}{Random seed} \\
                         &                                & 0        & 10       & 100       \\
\midrule
\multirow{3}{*}{Cornell}    & CAT                            &\textbf{75.8$\pm$3.5}          &72.9$\pm$4.1          &69.0$\pm$2.0           \\
                         & CAT(random\_cluster)            &72.4$\pm$3.2          &74.1$\pm$3.2          &70.4$\pm$1.7           \\
                         & CAT(high\_distraction)                   &73.3$\pm$3.0          &68.9$\pm$2.1          &\uwave{65.2$\pm$2.4}           \\
\multirow{3}{*}{Texas}    & CAT                            &65.2$\pm$2.4          &\textbf{70.2$\pm$2.2}          &66.6$\pm$3.4           \\
                         & CAT(random\_cluster)            &67.4$\pm$3.1          &69.5$\pm$2.1          &69.5$\pm$3.6           \\
                         & CAT(high\_distraction)                   &\uwave{61.9$\pm$2.9}          &68.4$\pm$3.1          &68.4$\pm$3.1           \\
\multirow{3}{*}{Wisconsin}    & CAT                            &62.5$\pm$4.3          &76.2$\pm$3.9          &\textbf{76.5$\pm$2.7}           \\
                         & CAT(random\_cluster)            &72.2$\pm$1.8          &73.0$\pm$1.9          &66.4$\pm$3.0           \\
                         & CAT(high\_distraction)                   &\uwave{60.9$\pm$2.2}          &68.9$\pm$2.0          &64.3$\pm$2.5           \\
\multirow{3}{*}{Chameleon}    & CAT                            &48.8$\pm$0.6          &\textbf{51.9$\pm$0.7}          &51.9$\pm$1.0           \\
                         & CAT(random\_cluster)            &48.5$\pm$0.5          &47.9$\pm$0.6          &48.8$\pm$1.1           \\
                         & CAT(high\_distraction)                   &45.9$\pm$0.8          &41.5$\pm$0.6          &\uwave{40.0$\pm$1.1}           \\
\multirow{3}{*}{Squirrel}    & CAT                            &\textbf{29.0$\pm$0.3}          &28.9$\pm$0.3          &28.4$\pm$0.3           \\
                         & CAT(random\_cluster)            &28.2$\pm$0.3          &27.1$\pm$1.6          &27.6$\pm$0.9           \\
                         & CAT(high\_distraction)                   &\uwave{24.5$\pm$2.7}          &27.3$\pm$1.9          &26.7$\pm$0.2          \\
\multirow{3}{*}{Actor}    & CAT                            &32.6$\pm$0.6          &\textbf{33.7$\pm$0.6}          &31.5$\pm$0.4           \\
                         & CAT(random\_cluster)            &32.9$\pm$0.5          &31.9$\pm$0.5          &31.7$\pm$0.5           \\
                         & CAT(high\_distraction)                   &30.6$\pm$0.6          &30.9$\pm$0.4          &\uwave{29.7$\pm$0.4}           \\
\multirow{3}{*}{Roman-empire}    & CAT                            &61.5$\pm$0.2          &\textbf{63.5$\pm$0.3 }         &62.2$\pm$0.2           \\
                         & CAT(random\_cluster)            &61.2$\pm$0.2          &61.5$\pm$0.2          &61.0$\pm$0.3           \\
                         & CAT(high\_distraction)                   &\uwave{48.8$\pm$0.2}          &51.0$\pm$0.3          &49.7$\pm$0.3           \\
\bottomrule
\end{tabular}
\end{table}

CAT consistently achieves the best performance, while CAT (high\_distraction) performs the worst. This comparison supports our Low distraction and High self-attention assumption and validates the efficacy of the Total Effect Estimation Module. CAT (random\_cluster) gets the medium performance, indicating the significance of the Class-level Semantic Clustering Module; to a certain extent, the comparison can also aid in quantifying the impact of each class on the performance of the model. In addition, it shows that the Total Effect Estimation Module makes a larger and more stable contribution to CAT's performance.

\begin{figure}
	\centering
	\includegraphics[width=5in]{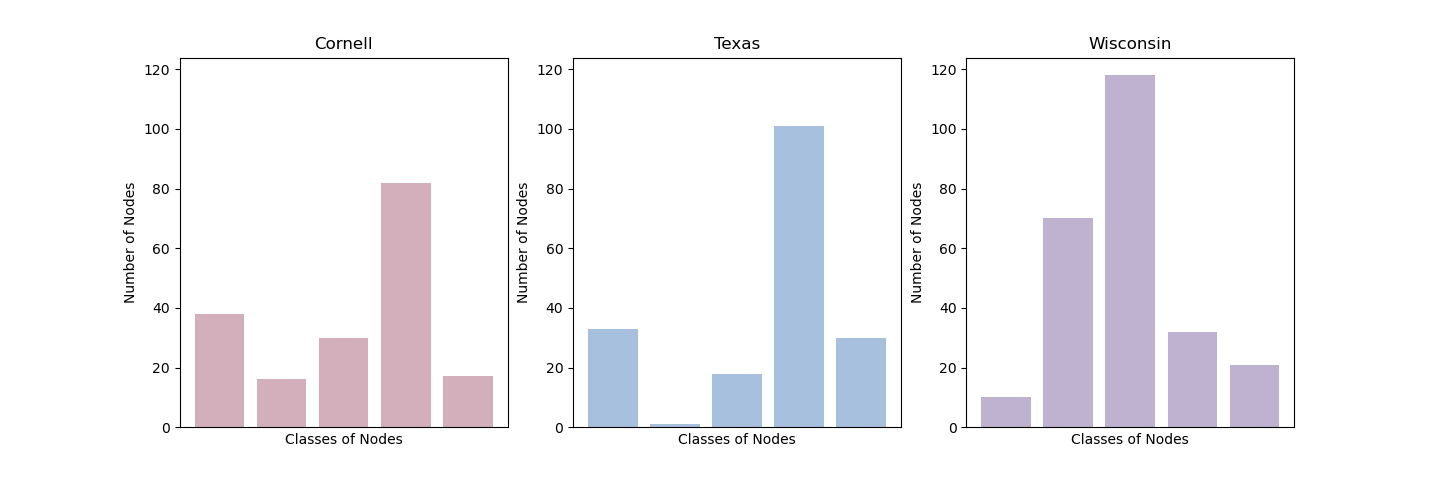}
	\caption{The class imbalance of small-size datasets.}
	\label{fig10}
\end{figure}

However, we also notice that CAT (random\_cluster) can achieve results comparable to or even exceeding those of CAT in very few cases. This suggests that the clustering results obtained by the Class-level Semantic Clustering Module need optimization, whereas random clusters perform better in some instances. This phenomenon is more striking on small-size datasets, possibly because of the class imbalance issues (as shown in Figure \ref{fig10}), which increases the clustering difficulty.

\subsubsection{Visualization\label{sec6.2.3}}
\begin{figure}
	\centering
	\subfloat[Cornell]{\includegraphics[width=.3\columnwidth]{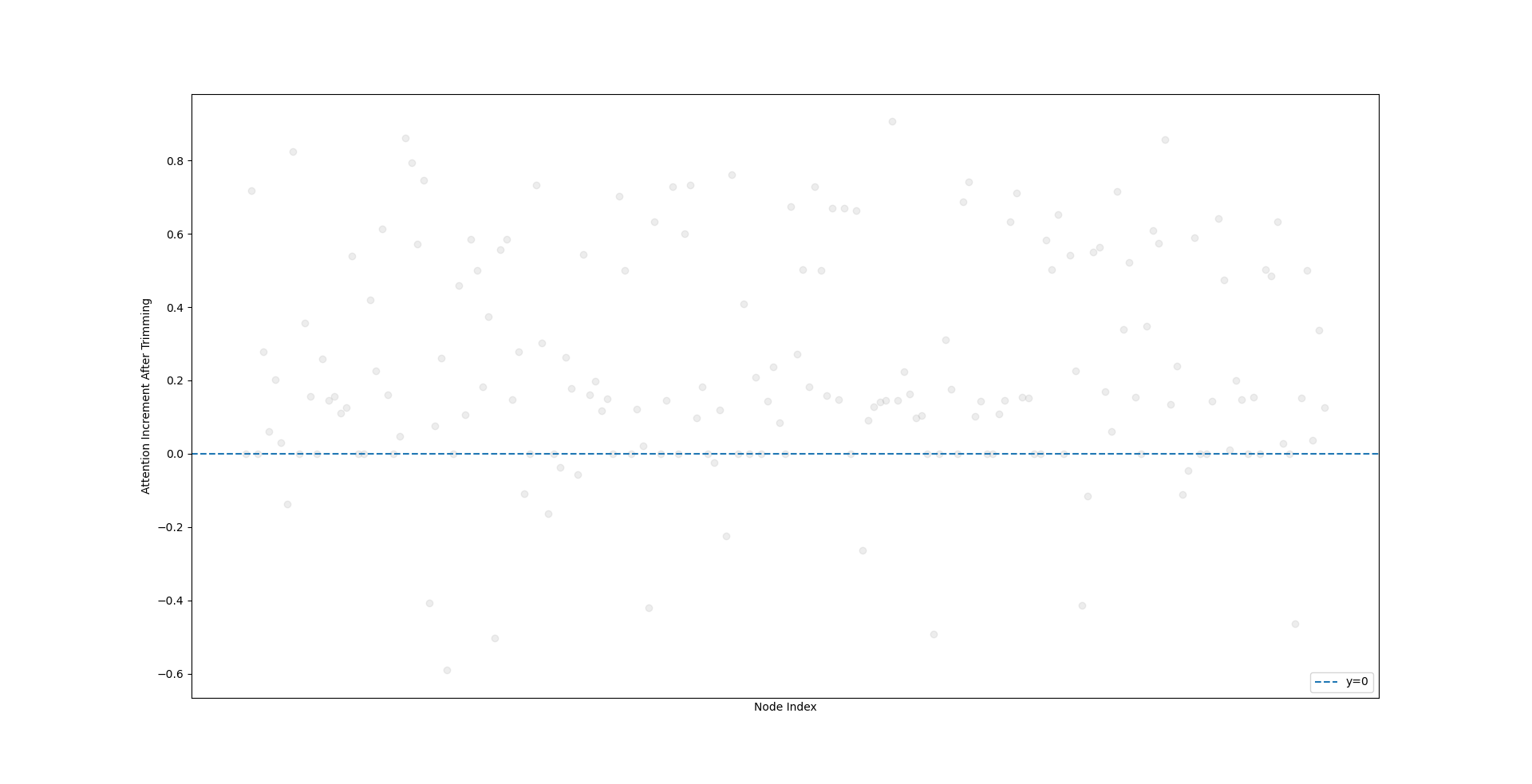}}
	\subfloat[Texas]{\includegraphics[width=.3\columnwidth]{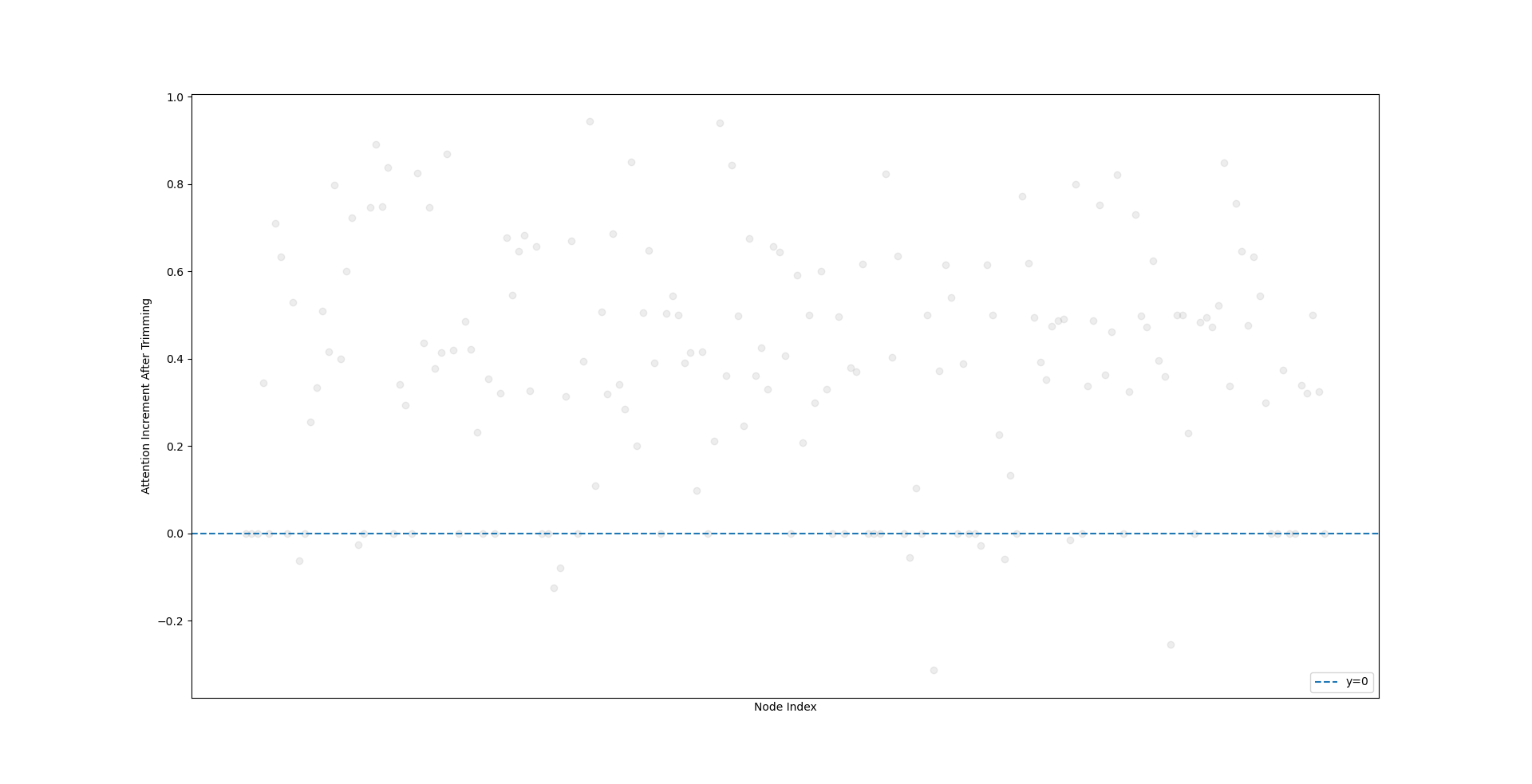}}
        \subfloat[Wisconsin]{\includegraphics[width=.3\columnwidth]{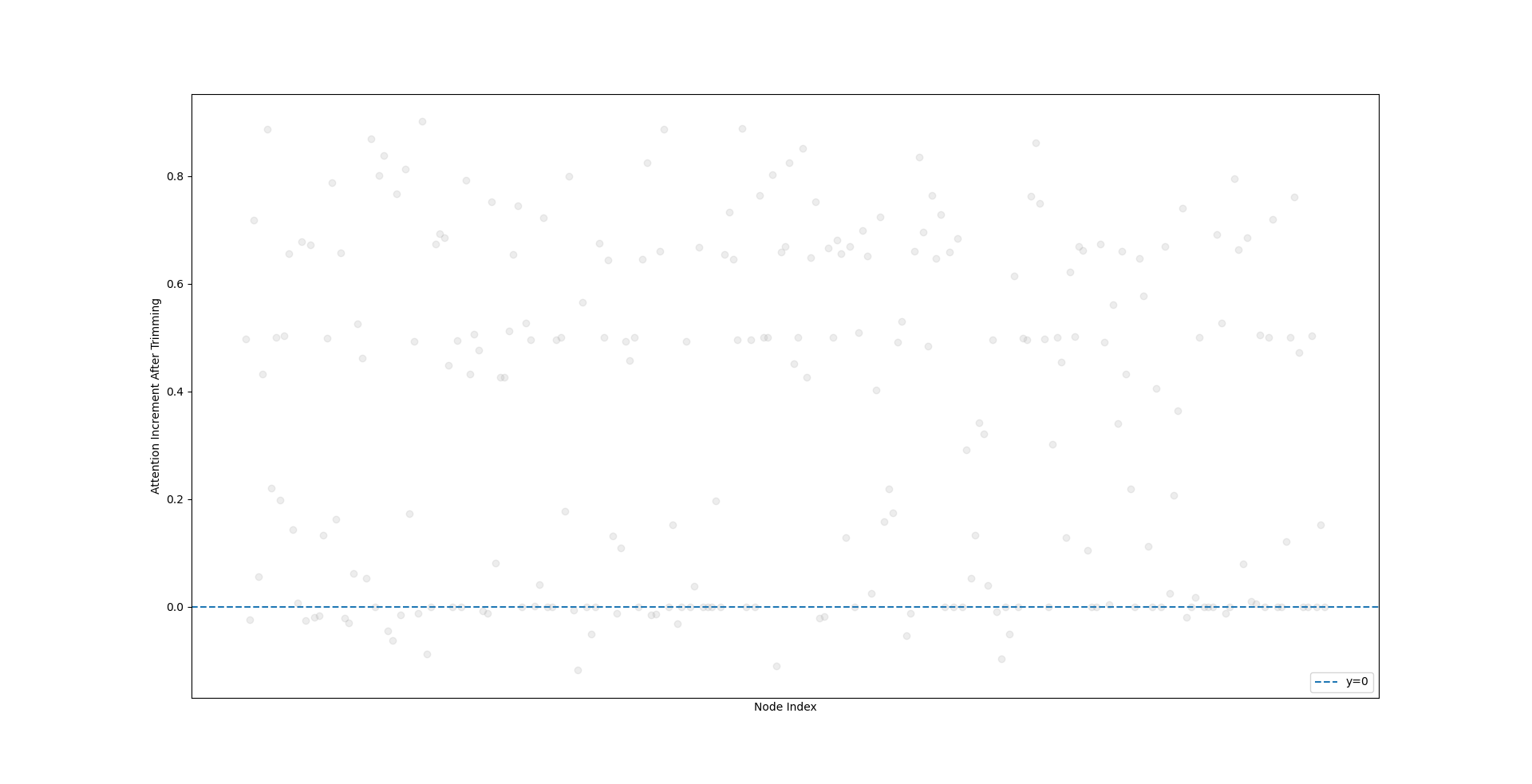}}\\
        \subfloat[Chameleon]{\includegraphics[width=.3\columnwidth]{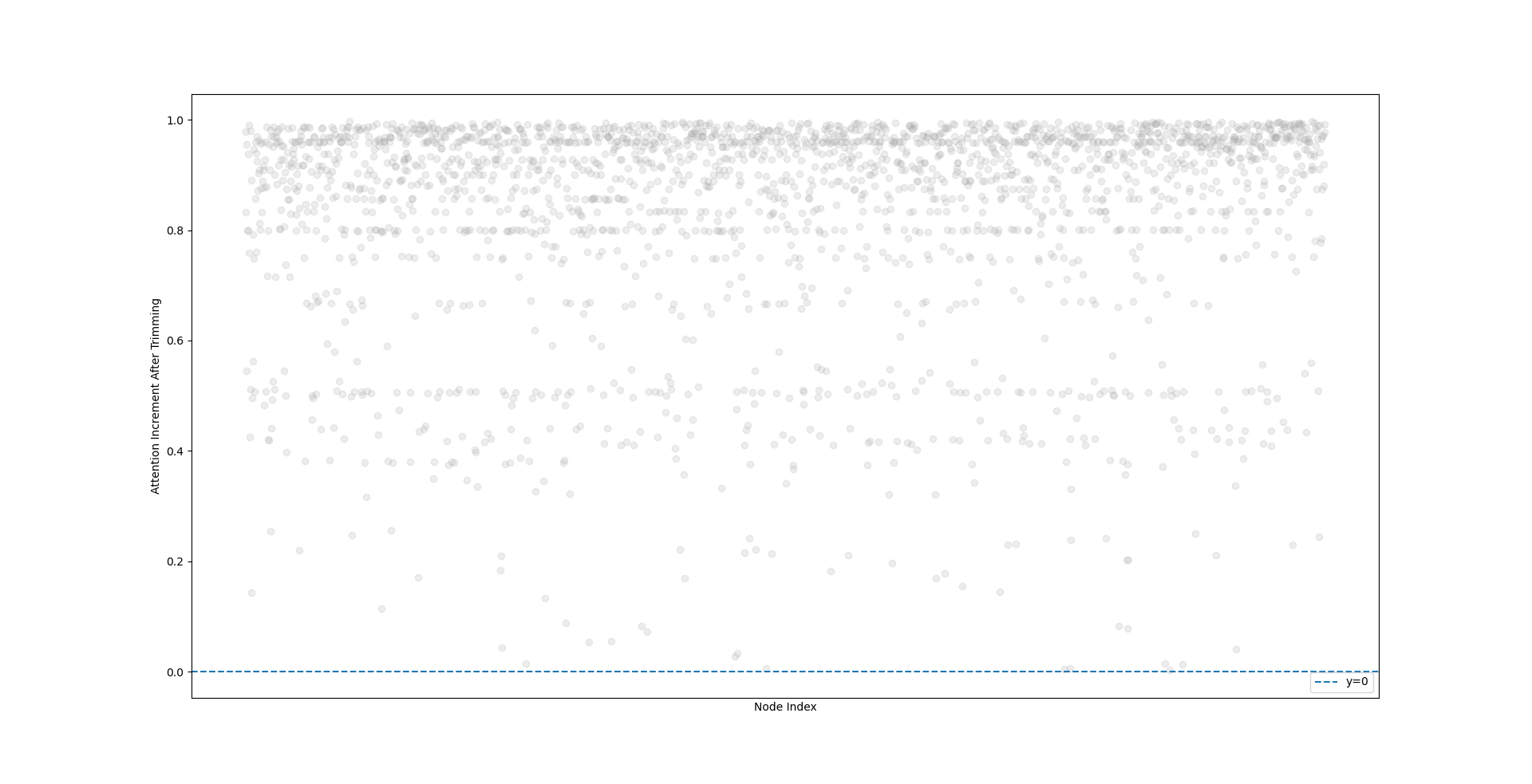}}
	\subfloat[Squirrel]{\includegraphics[width=.3\columnwidth]{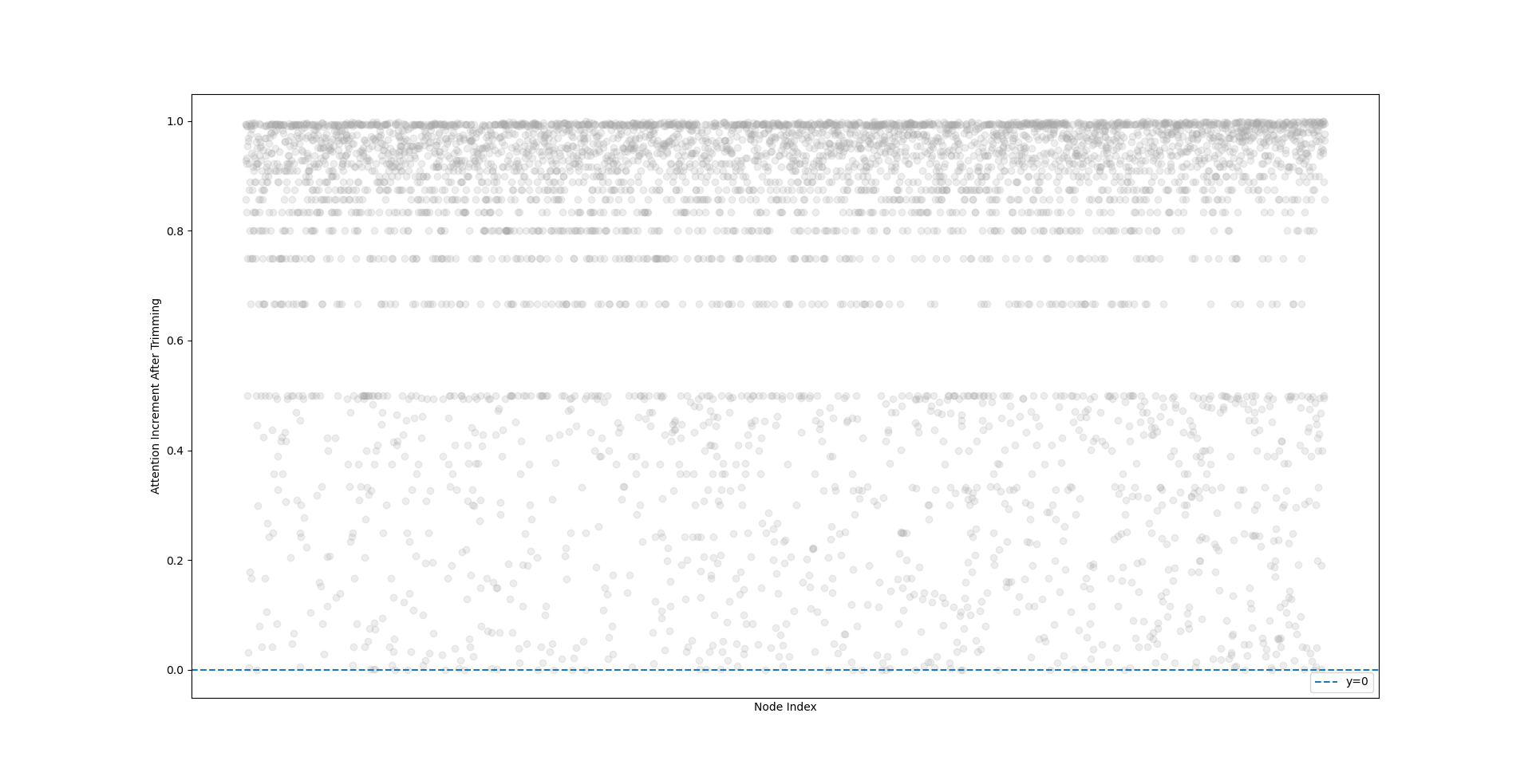}}
        \subfloat[Actor]{\includegraphics[width=.3\columnwidth]{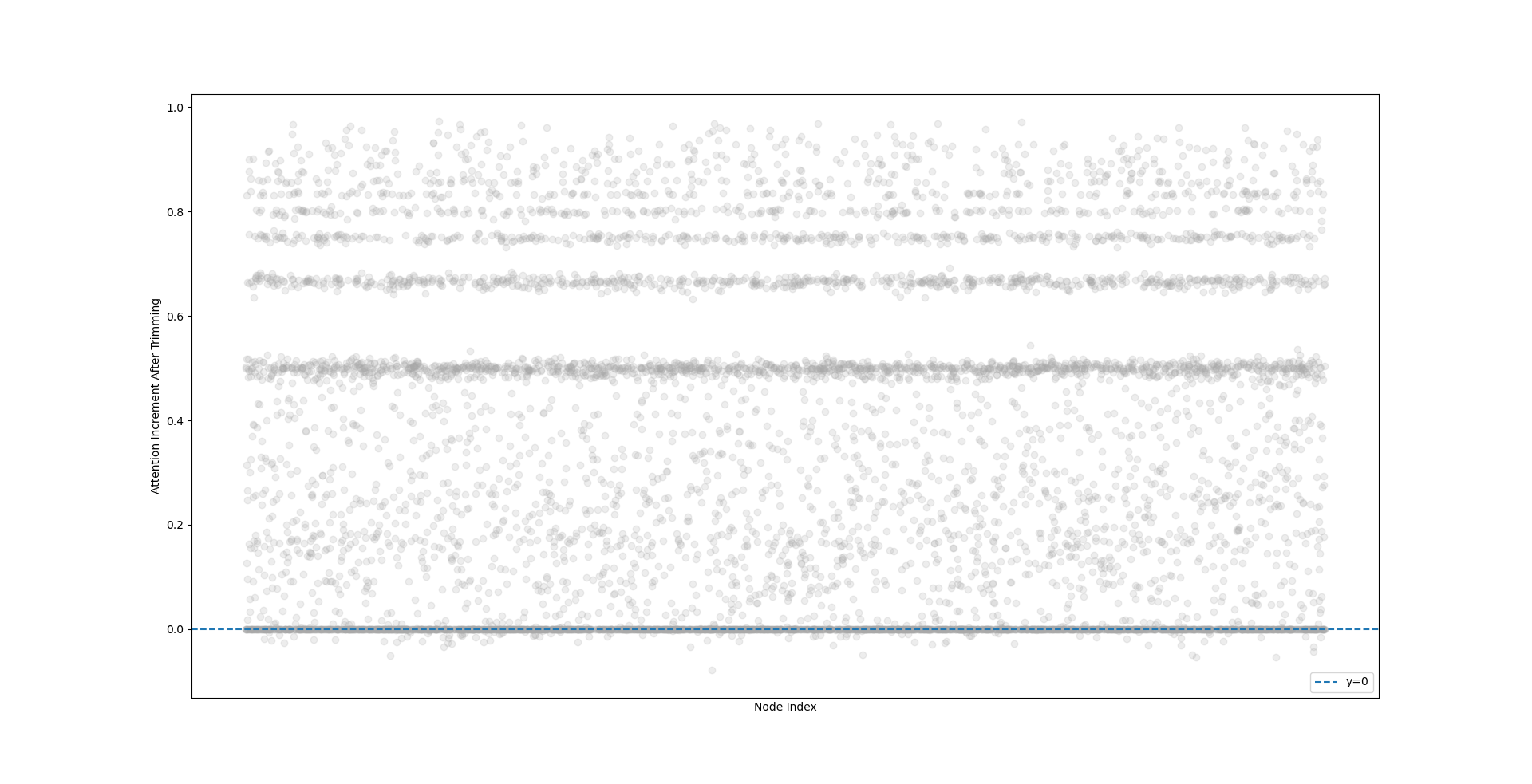}}\\
        \subfloat[Roman-empire]{\includegraphics[width=.3\columnwidth]{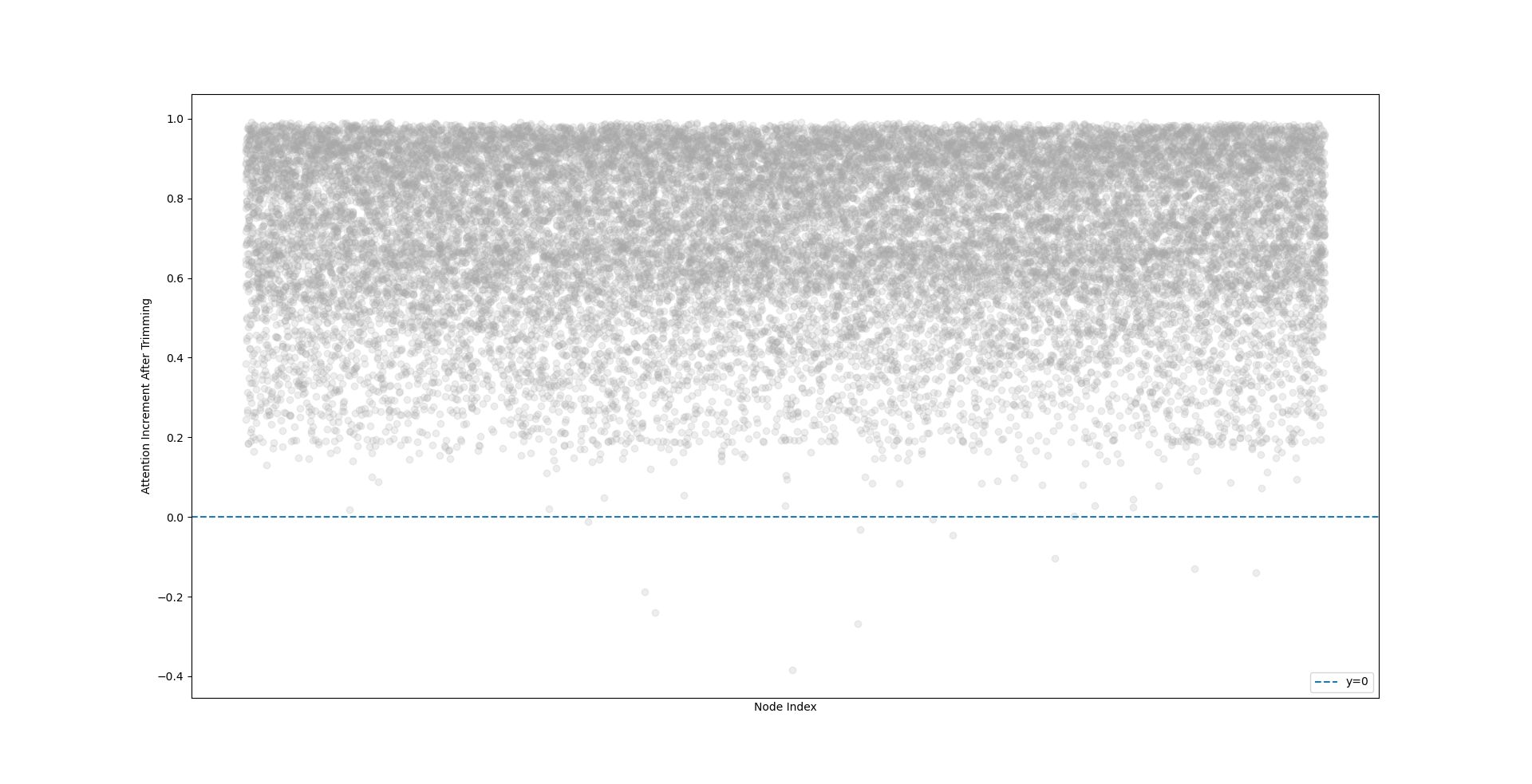}}
	\caption{Comparison between the self-attention values learned before and after graph trimming. The vertical coordinate represents the change in self-attention after graph trimming and a higher value represents a more significant enhancement of the self-attention level.}
        \label{fig11}
\end{figure}

\textbf{CAT can enhance the self-attention of central nodes}. To verify whether CAT enhances the central node’s self-attention and reduces the DE it suffers, we compare the final self-attention values learned by all nodes before and after trimming. We take CAT-unsup as an example and visualize the self-attention improvement after graph trimming in Figure \ref{fig11}. It can be observed that for the vast majority of nodes, the graph obtained by CAT can make the base GAT pay more attention to the nodes themselves and alleviate the neighbors’ distraction; while very few nodes exhibit decreased self-attention, possibly because the nodes already obtained high self-attention before trimming and their neighbors received more attention after trimming due to the reduction in the number of competitors. Fortunately, this situation is rare and does not affect the overall discrimination ability of the model.

\begin{figure}
	\centering
	\includegraphics[width=4in]{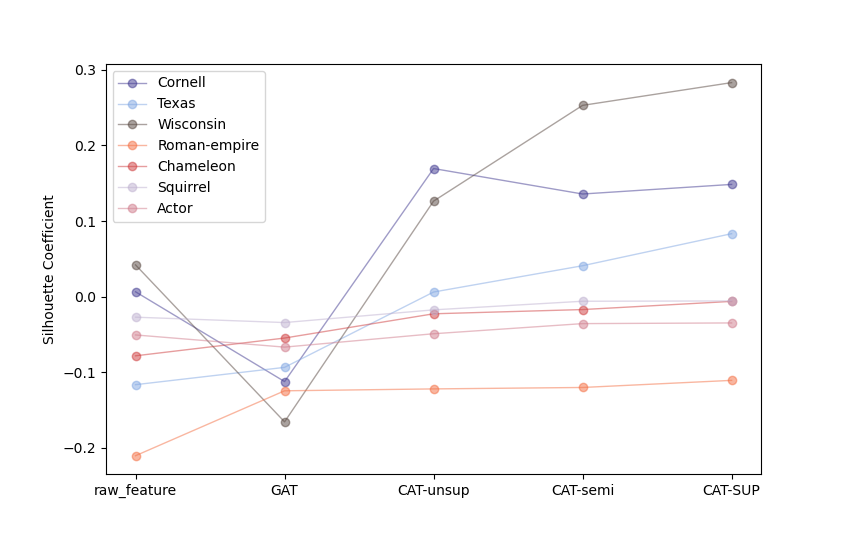}
	\caption{The silhouette coefficient (SC) of the learned node representations. The representations obtained by CAT  consistently achieve higher SCs than do both the original features and GAT representations.}
	\label{fig12}
\end{figure}

\textbf{CAT can alleviate the degradation of discrimination ability exhibited by the GAT}. To visualize whether the model's discrimination ability is improved, we conduct dimensionality reduction on the learned node representations and calculate their corresponding silhouette coefficient (SC). As shown in Figure \ref{fig12}, we use t-SNE to reduce the representations to two dimensions, where a higher SC represents an enhanced ability to discriminate between different classes. We compare the original input features, the representations output by base GATs, the representations output by CAT variants, and their corresponding SCs. The parameter settings yielding the highest node classification accuracy are selected as the representative result. We observe that the representations obtained by GAT involve a lower SC compared to that of the original features, indicating the discrimination ability degradation exhibited by GAT. In contrast, the representations obtained by CAT variants consistently achieve the highest SC, which implies that CAT can alleviate the discrimination ability degradation. The discrimination abilities of the three CAT variants are relatively close. Generally, CAT-sup has the highest discrimination capability, followed by CAT-semi, with CAT-unsup performing the worst. Although the SC obtained by our method is not sufficiently high, it is adequate for achieving some improvement in mitigating the decrease in discrimination ability caused by LAMP.

\begin{figure}
	\centering
	\subfloat[Chameleon]{\includegraphics[width=.5\columnwidth]{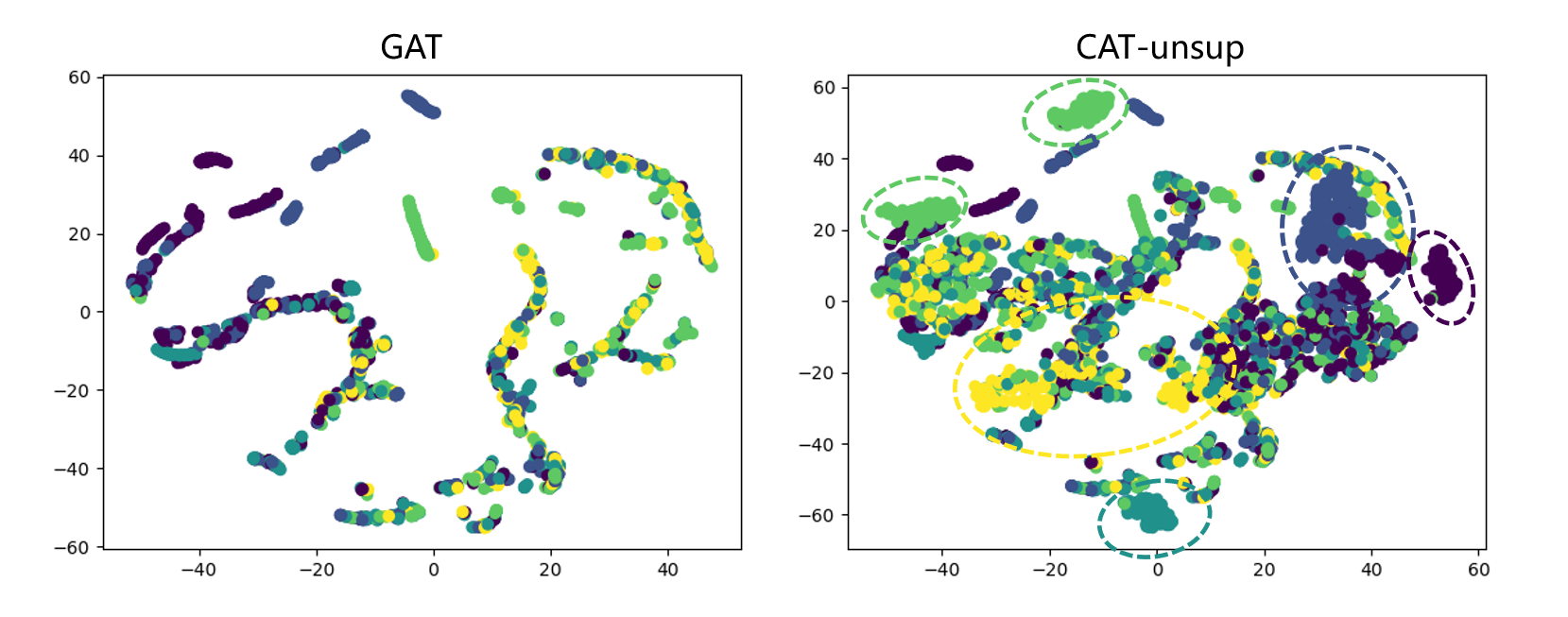}}
	\subfloat[Cornell]{\includegraphics[width=.5\columnwidth]{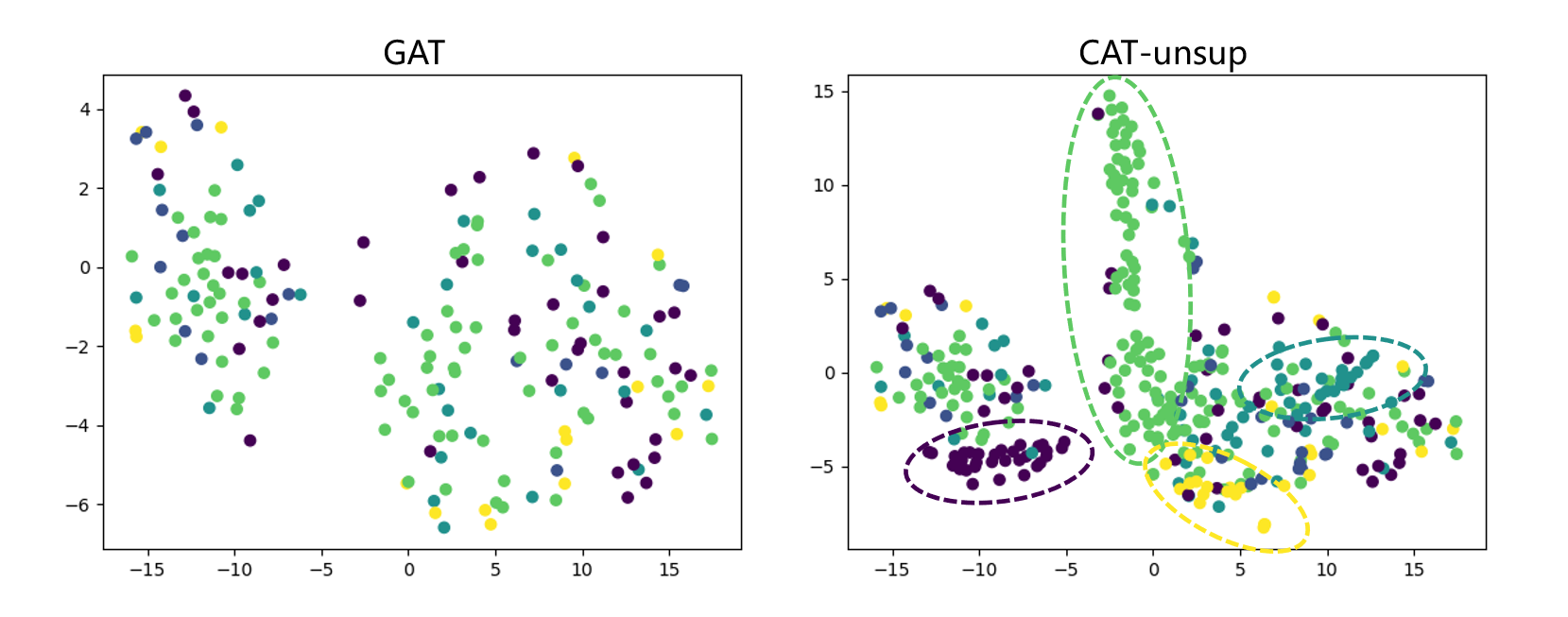}}
	\caption{Visualization of the embeddings in the representation space learned by GAT and CAT-unsup.}
        \label{fig1314}
\end{figure}

\begin{figure}
	\centering
	\includegraphics[width=6in]{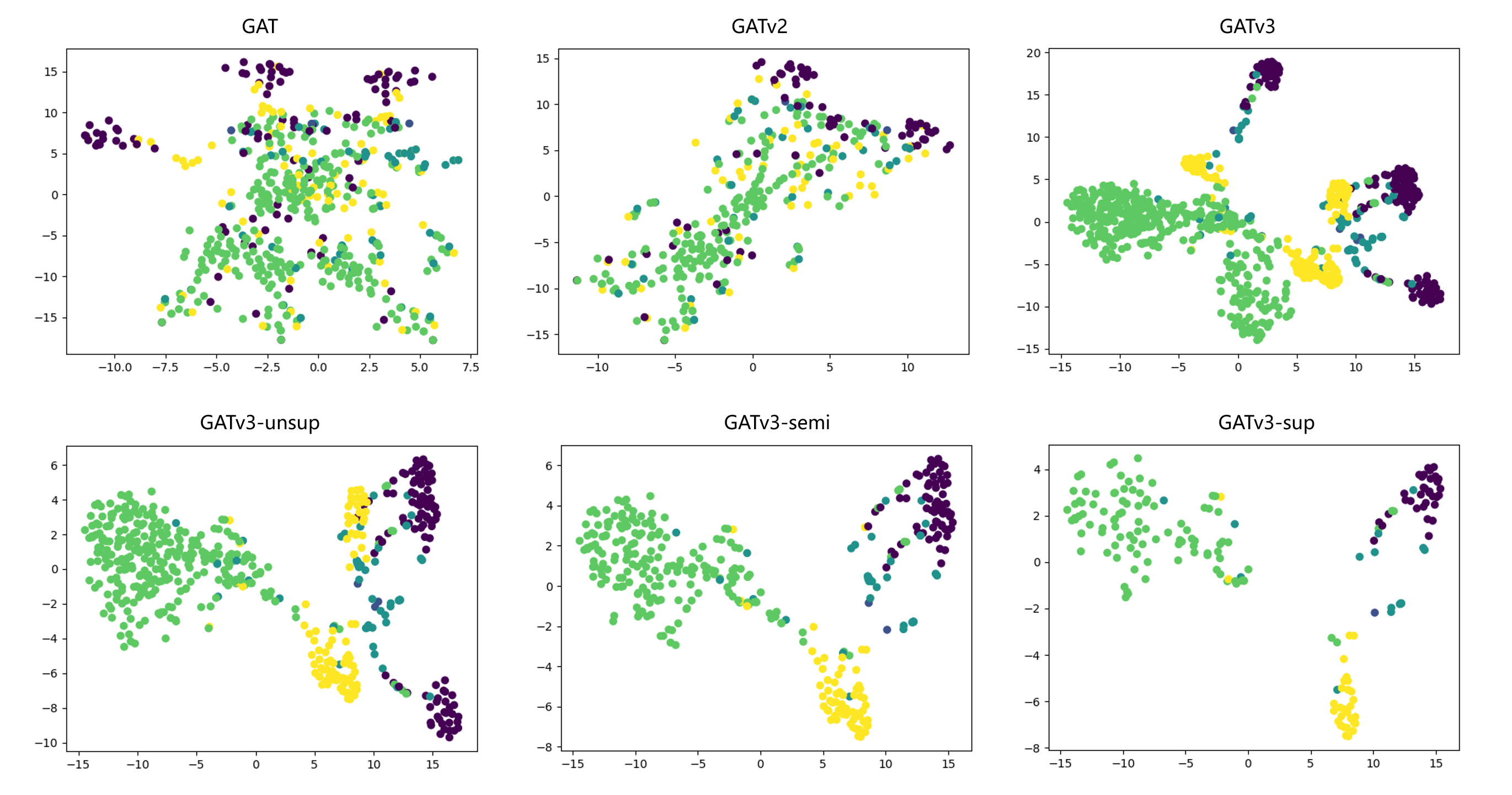}
	\caption{Visualization of the embeddings of the Texas dataset learned by different base GATs and variants of CATv3.}
	\label{compare}
\end{figure}

\begin{figure}
	\centering
	\includegraphics[width=5in]{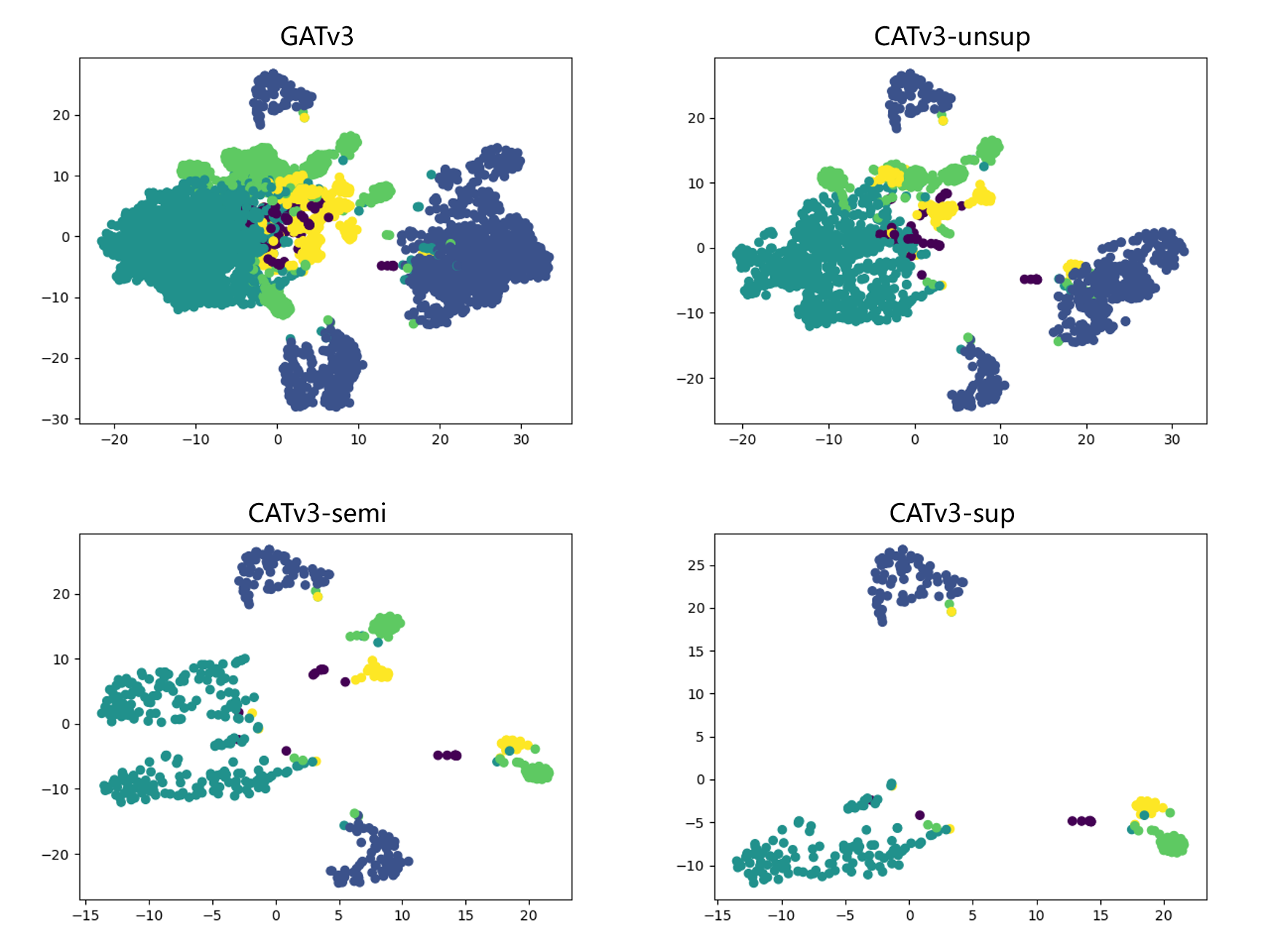}
	\caption{Visualization of the embeddings of the Wisconsin dataset learned by GATv3, GATv3-unsup, GATv3-semi and GATv3-sup.}
	\label{figadd3}
\end{figure}

\textbf{CAT can embed graphs to a representation space approaching the ideal semantic space}. We observe that more nodes learned by CAT have significant clustering tendencies compared to GAT, which is manifested as more clustered structures in visualized figures. As shown in Figure \ref{fig1314}, on the Chameleon dataset, CAT can identify more clusters than GAT such as the \textcolor[RGB]{71,143,160}{dark green} and \textcolor[RGB]{62,7,81}{dark purple} clusters. On the Cornell dataset, the representations obtained by CAT bring nodes belonging to the same class closer in the representation space such as the \textcolor[RGB]{123,198,110}{light green} and \textcolor[RGB]{62,7,81}{dark purple} clusters, implying that the nodes are located closer to the cluster center and are easier to distinguish from the clusters in other classes. For different base models, Figure \ref{compare} exhibits a slight difference between GAT and GATv2, while GATv3 which is specifically designed for handling heterophilic graphs, learns more distinguishable representations. Nevertheless, CATv3-unsup is capable of learning more compact clusters compared to GATv3 such as the \textcolor[RGB]{123,198,110}{light green} clusters. CATv3-semi and CATv3-sup can further learn superior representations. As shown in Figure \ref{figadd3}, there is an evident trend that with more Class-level Semantic Cluster information, CAT variants can learn more compact and separable clusters. As the base model, GATv3 learns the Semantic Cluster distribution with the lowest cluster cohesion and separation. The distances between the clusters learned by CATv3-sup are maximized, and the nodes within a cluster are closest to the cluster center, while CATV3-unsup exhibits the opposite performance. This phenomenon highlights the significance of Class-level Semantic Clusters.



\section{Discussions\label{sec7}}

\subsection{Fundamental Hypothesis on Heterophilic Graphs} \label{sec7.1}
\begin{figure}[htbp]
\centering
\includegraphics[width=3in]{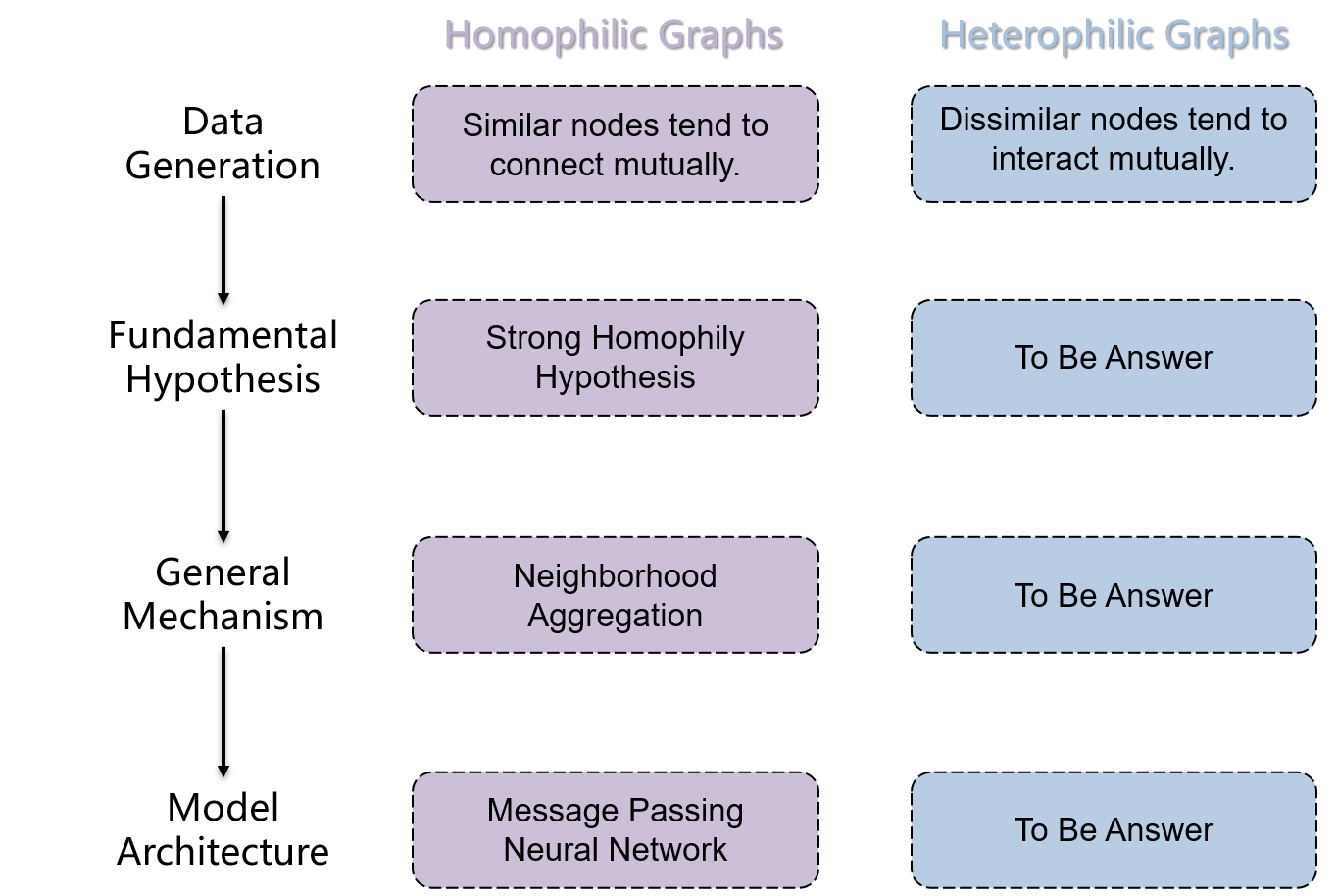}
\caption{Discussion regarding heterophilic graphs.}
\label{fig15}
\end{figure}

A fundamental issue behind improving the performance of GNNs on heterophilic graphs is the assumption about the generation mechanism of heterophilic graphs. The strong homophily hypothesis holds that connections between nodes are generated because they are sufficiently similar, thus deriving a neighboring aggregation mechanism, which the heterophilic graphs don’t hold. This raised an important question for heterophilic graphs, which we depict in Figure \ref{fig15}.

\textbf{Question: What is the fundamental hypothesis underlying heterophilic graphs?} How to build a brand-new graph representation learning mechanism for heterophilic graphs? It requires us to propose new inductive biases based on the generation mechanism of heterophilic graphs. This is a challenging, landmark mission.

\subsection{Limitations of CAT and Future Works} \label{sec7.2}

\textbf{The lack of general hypothesis for heterophilic graphs}. In this paper, we hypothesized that the generation mechanism underlying heterophilic graphs will derive models different from the current neighboring aggregation models. Based on this insight, we offered a possible way, and have made a preliminary attempt on GATs: \textbf{to make the node concentrate more on itself instead of relying excessively on all neighbors.} Specifically, we employ causal inference methods to identify those neighbors that can help central nodes concentrate on themselves as much as possible. Our solution relies on the attention mechanism of GATs, which is not a generalized solution. Determining how to derive a general heterophilic graph representation learning framework is an endeavor for the future.

\textbf{The lack of an effective way to learn optimal class-level Semantic Cluster}. According to our \hyperref[Hyp1]{Class-level Semantic Space Hypothesis}, the ideal semantic space is compact and separable. Considering the semi-supervised learning paradigm of node classification tasks, it is more reasonable for the Class-level Semantic Clustering Module to adopt an unsupervised or semi-supervised manner. The challenges concern high dimensionality, sparsity, and low semantic expressiveness of original node features. In the future, it is imperative to explore more effective methods for learning a better Class-level Semantic Space with less label information, including unsupervised, semi-supervised, and self-supervised learning methods. Training self-adaption modules is also explorable. 

\textbf{The lack of extension for transformer-based graph learning methods}. We only investigate the discrimination ability degradation of GNNs when meeting heterophilic graphs caused by the LAMP mechanism. However, the transformer \cite{add6}, a neural network with a powerful global attention mechanism, can be transferred to graph learning tasks. Whether graph transformers \cite{update2} face the same challenges as GATs on heterophilic graphs, and how to extend the current strategy behind this work to the graph transformer architecture is worthy of future investigation. 

To comprehensively and visually assess the proposed method, we applied SWOT analysis \cite{update3} in CAT. More future endeavors can be inferred from the SWOT matrix (Table \ref{tab5}), such as base model reinforcement and extension, and high-quality heterophilic graph benchmarks. The result clearly shows that the Distraction Effect and Distraction Neighbors identified in CAT have different practical implications in various scenarios and can be applied to analyzing real-world business datasets. For example, Distraction Neighbors may represent the different roles of friends in heterophilic social networks.


\begin{table}
\caption{SWOT matrix of CAT.}
\label{tab5}
\begin{tabular}{ll}
\toprule
\multicolumn{2}{l}{\textbf{Internal Factors}} \\
\midrule
\textbf{Strengths}             & \textbf{Weakness}            \\
\midrule
\makecell[l]{1. No need to alter the base GAT model. \\2. No need to seek for more similar neighbors .\\3. Plug-and-play and applicable to any\\ LAMP-driven GATs.\\4. Afford causal interpretation.}
& \makecell[l]{1. Performance relies on the discrimination ability of the\\ base GAT model.\\2. Performance relies on the label distribution of raw data.}         \\
\toprule
\multicolumn{2}{l}{\textbf{External Factors}} \\
\midrule
\textbf{Opportunities}             & \textbf{Threats}            \\
\midrule
\makecell[l]{1. High accuracy when label information is sufficient to\\ uncover the category distribution.\\2. High accuracy when an effective clustering method \\is implemented. \\ 3. Explain the role of nodes in real-world business\\ scenarios like social network user analysis. }          
& \makecell[l]{1. Worse performance when label information is insufficient.\\ 2. Worse performance when the adopted clustering method\\ performs poorly.\\ 3. Unavailable when the base GAT model fails to execute.}         \\
\bottomrule
\end{tabular}
\end{table}

\section{Conclusion\label{sec8}}
To cope with the significant degradation of node classification performance exhibited by GATs on heterophilic graphs, we propose a Causal graph Attention network for Trimming heterophilic graphs (CAT). Three representative GATs are employed as the base model and their discrimination ability can be significantly improved after adopting CAT. Specifically, we propose a new hypothesis for GATs on heterophilic graphs, Low Distraction and High Self-Attention, which suggests enabling the central node to concentrate on itself and reduce distraction from neighbors. Based on this hypothesis, we leverage causal inference methods to estimate Distraction Effect and identify Distraction Neighbors. Distraction Neighbors are removed via graph trimming, allowing the base GAT model to achieve better node classification performance by maintaining self-attentions. Compared with existing methods, our method eliminates the need to alter the architecture of GATs or search for more neighbors globally; instead, it learns a new graph structure to obtain a better attention distribution. The experiments show that our method achieves significant performance improvements in node classification tasks on seven heterophilic graphs of three sizes. In addition, the framework of our method can be applied to any LAMP-driven model.

\section*{Acknowledgement}
This research was funded by the National Natural Science Foundation of China under Grant 42271481 and the Natural Science Foundation of Hunan Province under Grant 2022JJ30698. This work was carried out in part using computing resources at the High Performance Computing Platform of Central South University.

\bibliographystyle{unsrt}
\bibliography{reference}

\end{document}